\documentclass{article}
\PassOptionsToPackage{numbers,sort&compress}{natbib}
\usepackage[main, final]{neurips_2026}

\usepackage[utf8]{inputenc} 
\usepackage[T1]{fontenc}    
\usepackage{hyperref}       
\usepackage{url}            
\usepackage{booktabs}       
\usepackage{amsfonts}       
\usepackage{nicefrac}       
\usepackage{microtype}      
\usepackage{xcolor}         
\usepackage{amsmath}
\usepackage{subcaption}
\usepackage{wrapfig}
\usepackage{algorithm}
\usepackage{algorithm}
\usepackage{algpseudocode}
 
\title{Formatting Instructions For NeurIPS 2026}

\author{%
  Bui Trong Duc\thanks{Corresponding Author}\qquad Huynh Thi Thanh Binh \\
  School of Information and Communications Technology\\
  Hanoi University of Science and Technology, Hanoi, Vietnam\\
  \texttt{\{ducbt,~binhht\}@soict.hust.edu.vn}
}

\usepackage{booktabs}
\usepackage{multirow}
\usepackage{adjustbox}
\usepackage{makecell}
\usepackage{siunitx}

\newcommand{\best}[1]{\textbf{#1}}
\usepackage{amsmath,amssymb,amsthm} %
\usepackage{algorithm} %
\usepackage{algpseudocode} 
\usepackage{tikz}
\usetikzlibrary{arrows.meta, positioning, fit}
\newtheorem{assumption}{Assumption}

\newtheorem{lemma}{Lemma}
\newtheorem{theorem}{Theorem}

\newtheorem{proposition}{Proposition}

\newcommand{\algname}{BECON}
\title{\algname: Belief-Conditioned Constrained Multi-Objective Reinforcement Learning under Drifting Preferences and Budgets}

\begin{document}

\maketitle

\begin{abstract}
Constrained preference-conditioned MORL typically assumes full task observability, treating preferences and safety budgets as stationary task signals. In real-world deployment, however, these signals are often unreliable proxies: intent may drift, safety tolerances may shift with context, and exposed descriptors may be noisy, delayed, or incomplete. We therefore formulate constrained preference-conditioned MORL as a partially observed task-specification problem, where the agent must navigate a fundamental conflict between unreliable external signals and latent interaction dynamics. We propose \textbf{\algname}, a belief-conditioned framework for constrained MORL with drifting preferences and budgets, which infers latent task context from recent histories to estimate preferences and safety budgets. These estimates are fused with the exposed task signal via uncertainty-dependent gates, enabling the policy to defer to observations under high epistemic uncertainty while correcting them when history is informative. The fused context jointly conditions preference optimization and context-adaptive constraint enforcement. Experiments on diverse multi-objective control tasks demonstrate that \algname\ consistently improves over state-of-the-art MORL baselines, achieving stronger robustness and constraint satisfaction under task drift.
\end{abstract}

\section{Introduction}
Real-world decision-making rarely reduces to optimizing a single stationary
objective \cite{zhao2025multi,sun2025aerial}. Autonomous vehicles must balance speed, comfort, energy use, and
safety; robotic assistants must trade off task efficiency, user preference, and risk; and resource allocation systems must adapt to evolving utility and feasibility requirements \cite{felten2024multi,van2026single,phan2025pareto}. Multi-objective reinforcement learning (MORL) provides a fundamental framework for
these settings by modeling returns as objective vectors and resolving trade-offs through preferences \cite{lu2023multiobjective,mu2025preference,yuan2025moduli}. To avoid training a separate policy for each trade-off, preference-conditioned MORL learns a single policy that can adapt to different preferences at deployment
\cite{basaklar2023pdmorl,li2026cola}. In parallel, constrained RL and budgeted Markov decision processes (MDPs)
address safety and resource requirements by imposing cost constraints during
policy optimization \cite{carrara2019budgeted}.

A key limitation of existing paradigms is the direct observability assumption, which regards the runtime task descriptor as a faithful proxy for the underlying objective \cite{ishfaq2024offline,liu2025efficient}. In
preference-conditioned MORL, this assumption often manifests by equating the observed preference vector with the intended utility trade-off \cite{li2024continual}. In constrained RL, the provided budget is
treated as the relevant feasibility threshold, with violations controlled by a global or weakly contextualized Lagrange multiplier \cite{guo2025constraint}. This assumption is fragile in deployed systems, where task descriptors are neither guaranteed to be stationary nor fully reliable. User preferences may drift within an episode as goals and priorities evolve; admissible safety tolerance may depend on recent interaction history, accumulated constraint violations, or context-specific feasibility; and exposed task contexts may be stale, noisy, delayed, or only partially specified \cite{cao2026learning}.

These factors induce a signal-reality gap, in which the runtime descriptor
available to the agent may no longer reflect the latent task specification that
determines the desired utility trade-off and admissible risk level. An agent that
conditions only on this descriptor can therefore optimize stale preferences,
miss newly emerging feasibility constraints, or apply safety penalties that are
inappropriate for the current context. We therefore consider constrained MORL under task drift as a partially observed
task-specification problem, where the hidden
context determines both the effective objective trade-off and the admissible cost
level. Recent reward and cost patterns provide indirect evidence about this
context, but such evidence can be ambiguous, especially over short histories.
Thus, the agent must not only infer latent task context from interaction history,
but also calibrate how much this inferred context should override or defer to the
exposed task signal. 

We introduce \algname, a belief-conditioned framework for constrained MORL under
drifting task specifications. Rather than assuming that runtime preferences and
budgets are directly observable task variables, \algname\ models them as noisy
manifestations of a latent task context. A variational belief over this context
is inferred from recent interactions and used to arbitrate, according to
uncertainty, between exposed descriptors and history-derived evidence. This
design goes beyond standard history conditioning: history is not simply added as
extra input, but is used to decide whether the current task signal should be
trusted or corrected. The resulting context conditions both multi-objective
policy optimization and adaptive constraint enforcement, enabling safety
penalties to vary with inferred feasibility rather than remain fixed across
deployment conditions.

Our contributions are as follows:
\begin{itemize}
    \item We formulate constrained MORL with drifting preferences and budgets as
    a partially observed task-specification problem, where exposed preference and
    budget signals are unreliable observations of the operative task.

    \item We develop a belief-conditioned constrained MORL algorithm that infers
    latent intent and context-dependent feasibility from recent interaction
    history, and fuses inferred and exposed task signals through
    uncertainty-dependent arbitration.

    \item Experiments across diverse constrained MORL benchmarks, supported by
    ablations and theoretical analysis, show that \algname\ improves adaptation,
    preference-aligned performance, and constraint satisfaction under task drift.
\end{itemize}

\section{Related Work}
MORL addresses sequential decision
making with vector-valued returns, where an agent must balance competing
objectives rather than optimize a single scalar reward. Linear scalarization remains a standard formulation because this approach reduces MORL to preference-indexed single-objective control while preserving access to
Pareto-optimal solutions under suitable conditions \cite{lu2023multiobjective}. However, solving a separate policy for
each preference is inefficient, motivating preference-conditioned and universal
policy methods that take preferences as inputs and generalize across the
preference simplex \cite{basaklar2023pdmorl}. Recent advances further improve Pareto-front coverage and controllability by discovering constrained Pareto sets \cite{liu2025efficient}, sharing latent knowledge across preferences and enabling preference-controllable optimization
\cite{pmlr-v267-yang25ax}, and modeling the distribution of multi-objective
returns beyond their expectations \cite{NEURIPS2023_32285dd1}. A complementary line studies continual MORL and latent preference inference, aiming to improve retention, transfer, or recovery of preference-conditioned behavior from incomplete task information \cite{li2024continual,cao2026learning}.
These works highlight the difficulty of relying solely on explicit preferences,
but they do not jointly model preference and budget signals as uncertain runtime
observations in constrained MORL. In parallel, constrained and safe RL formalize safety, resource, or risk requirements through budgeted MDPs, and primal-dual or Lagrangian optimization, where
constraint satisfaction is typically enforced by cost thresholds or adaptive
multipliers \cite{carrara2019budgeted,ICLR2024_2f8b5654}.Recent progress further sharpens constrained RL through reward-free exploration
\cite{pmlr-v162-miryoosefi22a}, approachability-based constraint satisfaction
\cite{jiang2024achieving}, adaptive resource allocation \cite{chemingui2026online}, conservative off-policy learning \cite{li2026offpolicy}, and constraint-conditioned policy selection under
varying safety limits \cite{guo2025constraint,chemingui2025constraint}. When safety constraints are coupled with multiple objectives, constrained MORL
offers a unified formalism for balancing utility trade-offs against admissible
risk \cite{ICLR2025_59ddfff7}. Yet most existing methods still assume that the
preference vector and safety budget are available as reliable task variables,
rather than noisy observations of a latent decision context. Orthogonal progress
in representation learning, non-stationary RL, and in-context RL suggests that
hidden task structure can be extracted from temporal prediction
\cite{schwarzer2021dataefficient}, bisimulation metrics
\cite{zhang2021learning}, causal shift representations
\cite{pmlr-v235-zhang24ah}, local regularization for changing environments
\cite{hamadanian2025online}, or recent trajectories as implicit task context
\cite{pmlr-v267-ma25x}. Although these methods show that history can reveal hidden task structure, they
do not directly address constrained preference-conditioned MORL under task
specification ambiguity, where exposed preferences and budgets are imperfect
observations of the latent utility trade-off and admissible safety level. A more detailed discussion is deferred to {Appendix~\ref{app:related}}.

\section{Constrained MORL under Task Drift}
\label{sec:task_drift}

We formulate a structured non-stationary variant of constrained multi-objective
reinforcement learning. Consider a multi-objective Markov decision process
$\mathcal{M}=(\mathcal{S},\mathcal{A},P,\mathbf{r},c,\gamma)$, where
$\mathcal{S}$ and $\mathcal{A}$ are the state and action spaces,
$P(\mathbf{s}'\mid \mathbf{s},\mathbf{a})$ is the transition kernel,
$\mathbf{r}:\mathcal{S}\times\mathcal{A}\rightarrow\mathbb{R}^{K}$ is a vector
reward with $K$ objectives, $c:\mathcal{S}\times\mathcal{A}\rightarrow\mathbb{R}$
is a nonnegative stage cost, i.e., $c(\mathbf{s},\mathbf{a})\ge 0$, and
$\gamma\in[0,1)$ is the discount factor
\cite{carrara2019budgeted,ICLR2025_59ddfff7,pmlr-v162-miryoosefi22a}. The base environment is fixed, but the task specification may evolve over time
and may be only partially observed. This setting isolates task drift from
environment drift, so the agent need not relearn the transition dynamics but
must infer the current preference and feasibility context from the exposed task
signal and recent interaction history.

Let $
\Omega
:=
\{
\boldsymbol{\omega}\in\mathbb{R}^{K}
\mid
\omega_i\ge 0,\ \forall i\in\{1,\ldots,K\},
\ \|\boldsymbol{\omega}\|_1=1
\}
$ be the preference space.
At time $t$, the operative task is
$\tau_t=(\boldsymbol{\omega}_t,B_t)$, where
$\boldsymbol{\omega}_t\in\Omega$ is the current preference vector and
$B_t\in\mathcal{B}$ is the current cost budget. The preference determines the
scalarized reward, while the budget defines the feasible cost level. Given
$\mathbf{r}_t=\mathbf{r}(\mathbf{s}_t,\mathbf{a}_t)$ and
$c_t=c(\mathbf{s}_t,\mathbf{a}_t)$, the time-varying constrained objective is
\begin{equation}
\max_{\pi}
\;
\mathbb{E}_{\pi}
\left[
\sum_{t=0}^{T-1}
\gamma^t
\boldsymbol{\omega}_t^{\top}\mathbf{r}_t
\right]
\qquad
\mathrm{s.t.}
\qquad
\mathbb{E}_{\pi}
\left[
c_t
\mid
\mathbf{s}_t,H_t
\right]
\le B_t,
\quad \forall t .
\label{eq:task_drift_objective}
\end{equation}
where $T$ and $H_t$ denote the finite decision horizon and the recent interaction history
available at time $t$, respectively.  This paper uses per-step constraints because safety tolerance may drift
with the current task context, making the acceptable instantaneous cost
time-varying. Additionally, non-stationarity enters through the task specification, specifically
$\tau_t=(\boldsymbol{\omega}_t,B_t)$ may change over time. Therefore, task drift
changes what is desirable or feasible, whereas environment drift would instead
change the data-generating process, e.g.,
$P_t(\mathbf{s}'\mid\mathbf{s},\mathbf{a})
\neq
P_{t'}(\mathbf{s}'\mid\mathbf{s},\mathbf{a})$, or change the reward and cost functions.

The operative task $\tau_t$ is not assumed to be fully observable. Instead, the
agent receives an exposed task signal
$\widehat{\tau}_t=(\widehat{\boldsymbol{\omega}}_t,\widehat{B}_t)$, where
$\widehat{\boldsymbol{\omega}}_t$ and $\widehat{B}_t$ may be noisy, delayed, or
incomplete measurements of the operative preference and budget. Therefore, the exposed task signal should be treated as a partial observation of the current
decision context rather than as a complete task specification. Since the current state $\mathbf{s}_t$ and exposed task signal
$\widehat{\tau}_t$ may not be determined fully, the policy is conditioned on a finite interaction history
$H_t=(\xi_{t-L},\ldots,\xi_{t-1})$, where $L$ is the history length and $\xi_t
=
\left[
\mathbf{s}_t,\,
\mathbf{a}_t,\,
\mathbf{r}_t,\,
\widehat{\boldsymbol{\omega}}_t,\,
c_t,\,
\widehat{B}_t
\right]$ is the transition feature stored at time $t$. We instantiate task drift through interpolation as follows $\boldsymbol{\omega}_{t+1} =
\Pi_{\Omega}
\left(
(1-\rho_{\omega})\boldsymbol{\omega}_t
+
\rho_{\omega}\bar{\boldsymbol{\omega}}_t
\right)$ and $B_{t+1}= (1-\rho_B)B_t+\rho_B\bar{B}_t$,
where $\bar{\boldsymbol{\omega}}_t\sim\nu_{\omega}$, $\bar{B}_t\sim\nu_B$, $\Pi_{\Omega}$ projects onto the preference simplex, $\nu_{\omega}$ and $\nu_B$ are sampling
distributions for refreshed preferences and budgets, and $\rho_{\omega},\rho_B\in[0,1]$ control the drift rates. Under partial task observability, we consider history-conditioned policies of the form $\pi
\left(
\mathbf{a}_t
\mid
\mathbf{s}_t,
H_t,
\widehat{\boldsymbol{\omega}}_t,
\widehat{B}_t
\right)$.

\section{\algname~Framework}
\label{sec:method}
\begin{figure}
    \centering
    \includegraphics[width=\linewidth]{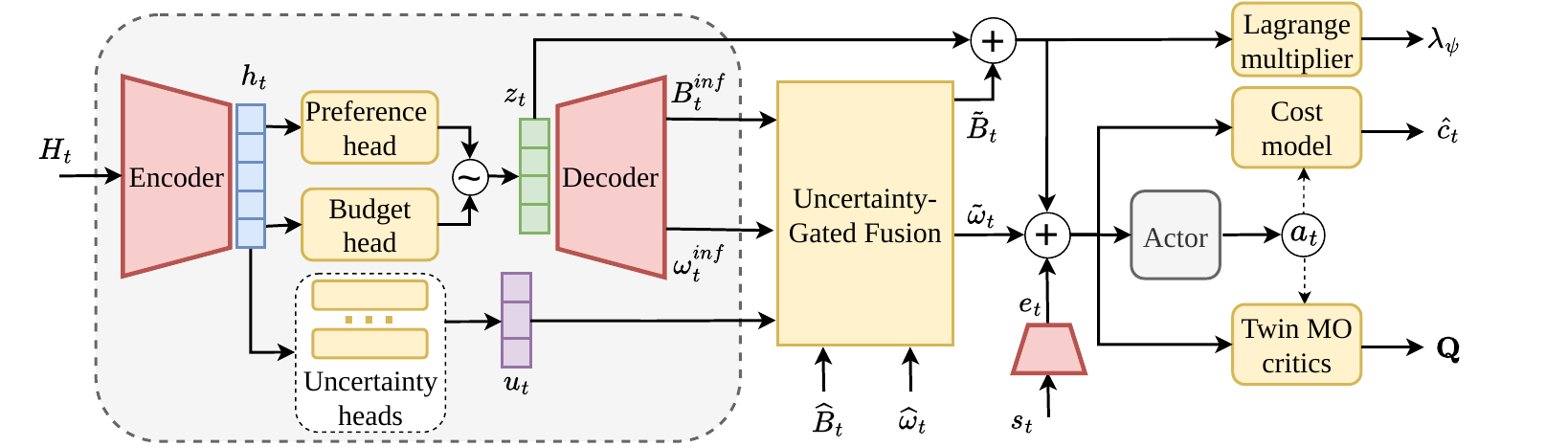}
 \caption{
Overview of \algname. A history encoder maps recent transitions $H_t$ to a
variational belief $\mathbf{z}_t$ over the latent task context. Uncertainty heads
gate the fusion between belief-derived estimates
$(\boldsymbol{\omega}^{\mathrm{inf}}_t,B^{\mathrm{inf}}_t)$ and the exposed
descriptor $\widehat{\tau}_t=(\widehat{\boldsymbol{\omega}}_t,\widehat{B}_t)$.
The fused context
$(\widetilde{\boldsymbol{\omega}}_t,\widetilde{B}_t)$ conditions the actor,
distributional critics, cost model, and context-dependent dual variable for
preference-aligned control with adaptive constraint enforcement.
}
    \label{fig:becon-overview}
\end{figure}
\algname\ implements the partially observed task-specification formulation
through a belief-conditioned constrained actor-critic architecture. At each time
step, a state encoder maps $\mathbf{s}_t$ to a latent state representation
$\mathbf{e}_t$, while a history encoder maps recent transition features $H_t$ to
a variational belief $\mathbf{z}_t$ over the unobserved task context. Rather
than treating the exposed descriptor
$(\widehat{\boldsymbol{\omega}}_t,\widehat{B}_t)$ as a ground-truth command,
\algname\ interprets it as a noisy observation and arbitrates between this signal
and the history-derived belief according to epistemic uncertainty. The resulting
fused context $(\widetilde{\boldsymbol{\omega}}_t,\widetilde{B}_t)$ conditions
both preference-aligned action selection and adaptive constraint enforcement.
Formally, the policy is written as $\pi_\theta
\left(
\mathbf{a}_t
\mid
\mathbf{s}_t,
H_t,
\widehat{\boldsymbol{\omega}}_t,
\widehat{B}_t
\right)
\equiv
\pi_\theta
\left(
\mathbf{a}_t
\mid
\mathbf{e}_t,
\mathbf{z}_t,
\widetilde{\boldsymbol{\omega}}_t,
\widetilde{B}_t
\right)$. Figure~\ref{fig:becon-overview} illustrates an overview of the proposed framework.

\subsection{History-Conditioned Belief and Task Decoding}
\label{subsec:belief-task-decoding}

The current state $\mathbf{s}_t$ and exposed task signal
$\widehat{\tau}_t=(\widehat{\boldsymbol{\omega}}_t,\widehat{B}_t)$ may not identify the operative
preference and feasibility context. We therefore encode the recent history $H_t$ into a stochastic
latent belief $\mathbf{z}_t\in\mathbb{R}^{d_z}$, where $d_z$ denotes the latent belief dimension. Specifically, a recurrent encoder produces
$\mathbf{h}_t=\mathrm{GRU}_\phi(H_t)$, followed by two heads
$\boldsymbol{\mu}_t=f_\phi^\mu(\mathbf{h}_t)$ and
$\log \boldsymbol{\sigma}_t^2=f_\phi^\sigma(\mathbf{h}_t)$, where
$\boldsymbol{\mu}_t,\boldsymbol{\sigma}_t\in\mathbb{R}^{d_z}$. The approximate posterior is $q_\phi(\mathbf{z}_t\mid H_t)
=
\mathcal{N}
\left(
\boldsymbol{\mu}_t,\;
\mathbf{\Sigma}_t
\right)$ with $\mathbf{\Sigma}_t
=
\mathrm{Diag}(\boldsymbol{\sigma}_t^2)$. A belief sample is drawn by reparameterization:
$\mathbf{z}_t=\boldsymbol{\mu}_t+\boldsymbol{\sigma}_t\odot\boldsymbol{\epsilon}$, where
$\odot$ denotes element-wise multiplication,
$\boldsymbol{\epsilon}\sim\mathcal{N}(\mathbf{0},\mathbf{I}_{d_z})$, and
$\mathbf{I}_{d_z}$ is the $d_z\times d_z$ identity matrix.

The latent belief $\mathbf{z}_t$ is intended to summarize hidden task context, including latent intent
and latent feasibility, rather than merely provide an auxiliary representation. To quantify uncertainty in this inference, we use the posterior standard
deviation as an aggregate uncertainty measure. Specifically, the scalar
belief uncertainty is computed from the posterior standard deviation
$\boldsymbol{\sigma}_t$ of $q_\phi(\mathbf{z}_t\mid H_t)$. In addition,
objective-specific log-variance heads applied to the hidden belief
representation produce per-objective uncertainty estimates
$\mathbf{u}_t=(u_{t,1},\ldots,u_{t,K})$. The scalar aggregate uncertainty and
the per-objective uncertainty estimates control how strongly \algname\ trusts
inferred task signals.
 
\algname\ decodes the belief $\mathbf{z}_t$ into two inferred task estimates:
an inferred preference
$\boldsymbol{\omega}^{\mathrm{inf}}_t\in\Omega$ and an inferred budget
$B^{\mathrm{inf}}_t\in\mathcal{B}$. Let $\mathbf{q}^{\omega}_t
=
g_{\omega}(\mathbf{z}_t)
\in\mathbb{R}^{K},$ and $
q^B_t
=
g_B(\mathbf{z}_t)
\in\mathbb{R}$ be the initial decoder outputs. They are mapped to valid task variables by
$\omega^{\mathrm{inf}}_{t,i}=
(\varphi(q^{\omega}_{t,i})+\varepsilon)/
\sum_{j=1}^{K}(\varphi(q^{\omega}_{t,j})+\varepsilon)$ and
$B^{\mathrm{inf}}_t=\varphi(q^B_t)$, where
$\varphi(x)=\log(1+\exp(x))$ is a positive activation and
$\varepsilon>0$ is a small constant. Thus,
$\boldsymbol{\omega}^{\mathrm{inf}}_t$ lies on the preference simplex and
$B^{\mathrm{inf}}_t$ is nonnegative. The decoded preference estimates the
current objective trade-off, while the decoded budget estimates the current
feasible cost level.

\subsection{Uncertainty-Gated Task Fusion}
\label{subsec:uncertainty-gated-fusion}

\algname\ uses belief uncertainty to control how much the policy should trust inferred task signals, specifically, between the exposed task signal
$\widehat{\tau}_t=(\widehat{\boldsymbol{\omega}}_t,\widehat{B}_t)$ and the
belief-derived estimate
$(\boldsymbol{\omega}^{\mathrm{inf}}_t,B^{\mathrm{inf}}_t)$. For nonempty histories, each objective gate is defined as $\alpha_{t,i}
= \alpha_{\max}\exp(-u_{t,i}),
\alpha_{\max}\in[0,1]$ and $\alpha_{t,i}=0$ when $H_t$ is empty. 
Let
$\boldsymbol{\alpha}_t=(\alpha_{t,1},\ldots,\alpha_{t,K})$. The fused preference and budget are
\begin{equation}
\widetilde{\boldsymbol{\omega}}_t
=
\Pi_{\Omega}
\left(
(1-\boldsymbol{\alpha}_t)\odot\widehat{\boldsymbol{\omega}}_t
+
\boldsymbol{\alpha}_t\odot
\boldsymbol{\omega}^{\mathrm{inf}}_t
\right), \quad \widetilde{B}_t
=
(1-\bar{\alpha}_t)\widehat{B}_t
+
\bar{\alpha}_t B^{\mathrm{inf}}_t
\label{eq:becon_fused_preference}
\end{equation}
where $\Pi_{\Omega}$ projects onto the preference simplex and $\bar{\alpha}_t=K^{-1}\sum_{i=1}^{K}\alpha_{t,i}$ is the average gate. Thus, uncertain beliefs defer to the exposed task signal, while confident beliefs can correct it using
history-derived context.

\subsection{Belief-Conditioned Constrained Actor--Critic}
\label{subsec:belief-conditioned-ac}
The current state is first mapped
to a latent embedding $\mathbf{e}_t=f_{\mathrm{enc}}(\mathbf{s}_t)$, which is shared across the actor,
critics, and cost model. The actor conditions on $\mathbf{e}_t$, the belief $\mathbf{z}_t$, the fused preference
$\widetilde{\boldsymbol{\omega}}_t$, and the fused budget $\widetilde{B}_t$:
$\pi_\theta(\mathbf{a}_t\mid \mathbf{e}_t,\mathbf{z}_t,
\widetilde{\boldsymbol{\omega}}_t,\widetilde{B}_t)$. For continuous control, it outputs Gaussian
parameters $(\boldsymbol{\mu}^{\pi}_t,\log\boldsymbol{\sigma}^{\pi}_t)
=
f_\pi
\left(
\mathbf{e}_t,
\mathbf{z}_t,
\widetilde{\boldsymbol{\omega}}_t,
\widetilde{B}_t
\right)$ and samples
$\tilde{\mathbf{a}}_t
=
\boldsymbol{\mu}^{\pi}_t
+
\boldsymbol{\sigma}^{\pi}_t
\odot
\boldsymbol{\epsilon}$,
where $\boldsymbol{\epsilon}\sim\mathcal{N}(\mathbf{0},\mathbf{I})$. The executed action is squashed as
$\mathbf{a}_t=\tanh(\tilde{\mathbf{a}}_t)$.

We use twin multi-objective critics
$\mathbf{Q}_j(\mathbf{e}_t,\mathbf{a}_t,\mathbf{z}_t,
\widetilde{\boldsymbol{\omega}}_t,\widetilde{B}_t)\in\mathbb{R}^{K}$ for $j\in\{1,2\}$. The
$i$-th component $Q_{j,i}$ estimates the value of objective $i$. Under the fused preference, the
scalarized critic is
\begin{equation}
Q_j^{\mathrm{scal}}
(\mathbf{s}_t,\mathbf{a}_t;\widetilde{\boldsymbol{\omega}}_t)
=
\sum_{i=1}^{K}
\widetilde{\omega}_{t,i}
Q_{j,i}
(\mathbf{e}_t,\mathbf{a}_t,\mathbf{z}_t,\widetilde{\boldsymbol{\omega}}_t,\widetilde{B}_t).
\label{eq:becon-scalq}
\end{equation}
In our implementation, each objective value is represented distributionally over atoms
$\mathcal{Z}=\{\zeta_1,\ldots,\zeta_N\}$ with $\zeta_n\in[V_{\min},V_{\max}]$. If $p_{j,i,n}$ is
the predicted probability of atom $\zeta_n$, then $Q_{j,i}=\sum_{n=1}^{N}p_{j,i,n}\zeta_n$.

The distributional cost model outputs $p^c_\nu$, and the predicted one-step cost is its expectation $\widehat{c}_t(\mathbf{a}_t)
=
\mathbb{E}_{p^c_\nu}[c]$.
Instead of using a single global Lagrange multiplier, which cannot adapt to time-varying feasibility requirements, \algname\ uses a context-dependent dual
variable
\begin{equation}
\lambda_\psi(\widetilde{B}_t,\mathbf{z}_t)
=
\varphi\!\left(f_\lambda(\widetilde{B}_t,\mathbf{z}_t)\right),
\label{eq:becon-dual}
\end{equation}
where $\varphi(x)=\log(1+\exp(x))$ ensures nonnegativity. This allows the safety penalty to vary
with the inferred feasibility context. Finally, the state embedding is regularized with a latent
transition model $\widehat{\mathbf{e}}_{t+1}=f_{\mathrm{dyn}}(\mathbf{e}_t,\mathbf{a}_t)$, encouraging
$\mathbf{e}_t$ to preserve predictive information about future states.

\subsection{Learning Objectives}
\label{subsec:becon-objectives}
All components are trained from mini-batches sampled from the replay buffer
$\mathcal{D}$. For each transition, we encode
$\mathbf{e}_{t+1}=f_{\mathrm{enc}}(\mathbf{s}_{t+1})$ and sample the next belief
$\mathbf{z}_{t+1}\sim q_\phi(\mathbf{z}_{t+1}\mid H_{t+1})$. 
The next action is
drawn under the stored fused task context $\mathbf{a}'_{t+1}$. Then, \algname{} selects the conservative twin-critic target, and applies the
standard distributional Bellman projection onto $\mathcal{Z}$ to obtain
$p_i^{\mathrm{tar}}$ for each objective $i$. The critic loss is
$\mathcal{L}_{\mathrm{critic}}
=
\sum_{j=1}^{2}
\mathbb{E}_{\mathcal{D}}
\left[
\sum_{i=1}^{K}
\widetilde{\omega}_{t,i}
\operatorname{CE}
\left(
p_i^{\mathrm{tar}},
p_{j,i}
\right)
\right],$ where $p_{j,i}$ is the predicted categorical distribution of critic $j$ for
objective $i$. Let $(\widehat{\mathbf{r}}^{\,b}_t,\widehat c^{\,b}_t,
\widehat{\boldsymbol{\omega}}^{\,\mathrm{rec}}_t,
\widehat B^{\,\mathrm{rec}}_t)
=
d_\eta(\mathbf{e}_t,\mathbf{a}_t,\mathbf{z}_t)$ denote auxiliary prediction heads from the belief representation, where
$\widehat{\mathbf{r}}^{\,b}_t$ and $\widehat c^{\,b}_t$ predict the observed
reward vector and cost, and
$(\widehat{\boldsymbol{\omega}}^{\,\mathrm{rec}}_t,\widehat B^{\,\mathrm{rec}}_t)$
reconstruct the exposed task descriptor. The belief objective is $\mathcal{L}_{\mathrm{belief}}
=
\lambda_{\mathrm{pred}}
\mathbb{E}_{\mathcal{D}}
\left[
\left\|
\widehat{\mathbf{r}}^{\,b}_t
-
\mathbf{r}_t
\right\|_2^2
+
\left(
\widehat c^{\,b}_t
-
c_t
\right)^2
\right]
+
\lambda_{\mathrm{obs}}
\mathbb{E}_{\mathcal{D}}
\left[
\left\|
\widehat{\boldsymbol{\omega}}^{\,\mathrm{rec}}_t
-
\widehat{\boldsymbol{\omega}}_t
\right\|_2^2
+
\left(
\widehat B^{\,\mathrm{rec}}_t
-
\widehat B_t
\right)^2
\right]
+
\beta_{\mathrm{KL}}
\mathbb{E}_{\mathcal{D}}
\left[
\operatorname{KL}
\left(
q_\phi(\mathbf{z}_t\mid H_t)
\,\|\,
\mathcal{N}(\mathbf{0},\mathbf{I}_{d_z})
\right)
\right].$
The dynamics and cost losses regularize the state embedding and train the
distributional cost model
\begin{equation}
\begin{aligned}
\mathcal{L}_{\mathrm{dyn}}
=
\mathbb{E}_{\mathcal{D}}
\left[
\left\|
f_{\mathrm{dyn}}(\mathbf{e}_t,\mathbf{a}_t)
-
\mathbf{e}_{t+1}^{-}
\right\|_2^2
\right], \quad
\mathcal{L}_{\mathrm{cost}}
=
\mathbb{E}_{\mathcal{D}}
\left[
\operatorname{CE}
\left(
p_t^{c,\mathrm{tar}},
p_t^c
\right)
\right],
\end{aligned}
\label{eq:becon-ldyn-lcost}
\end{equation}
where $\mathbf{e}_{t+1}^{-}$ is a stop-gradient target and $p_t^{c,\mathrm{tar}}$
is obtained by projecting the observed cost $c_t$. The actor minimizes an entropy-regularized constrained objective under the fused
task context
\begin{equation}
\mathcal{L}_{\mathrm{actor}}
=
\mathbb{E}_{\mathcal{D}}
\left[
\alpha_H
\log
\pi_\theta(\mathbf{a}_t\mid\cdot)
-
\min_{j\in\{1,2\}}
Q_j^{\mathrm{scal}}
\left(
\mathbf{s}_t,\mathbf{a}_t;
\widetilde{\boldsymbol{\omega}}_t
\right)
+
\kappa_t
\lambda_\psi(\widetilde B_t,\mathbf{z}_t)
\left[
\widehat c_t(\mathbf{a}_t)
-
\widetilde B_t
\right]_+
\right],
\label{eq:becon-lactor}
\end{equation}
where $\widehat c_t(\mathbf{a}_t)=\mathbb{E}_{p_t^c}[c]$ is the expected cost
from the distributional cost model, $\kappa_t\in[0,1]$ is the constraint warm-up
coefficient. The context-dependent dual variable is updated by minimizing $\mathcal{L}_{\lambda}
=
-
\mathbb{E}_{\mathcal{D}}
\left[
\lambda_\psi(\widetilde B_t,\mathbf{z}_t)
\left(
\widehat c_t(\mathbf{a}_t)
-
\widetilde B_t
\right)
\right].$ Since $\lambda_\psi$ is nonnegative, this update increases the penalty when the
predicted cost exceeds the fused budget and relaxes it otherwise. The auxiliary and value-learning objectives are optimized jointly as $\mathcal{L}_{\mathrm{main}}
=
\mathcal{L}_{\mathrm{critic}}
+
\lambda_{\mathrm{belief}}\mathcal{L}_{\mathrm{belief}}
+
\lambda_{\mathrm{dyn}}\mathcal{L}_{\mathrm{dyn}}
+
\lambda_{\mathrm{cost}}\mathcal{L}_{\mathrm{cost}}.$ Each iteration updates $\mathcal{L}_{\mathrm{main}}$, $\mathcal{L}_{\mathrm{actor}}$,
and $\mathcal{L}_{\lambda}$, followed by soft target-network updates.

\section{Theoretical Analysis}
\label{sec:theory}
Let $x_t=(\mathbf{s}_t,H_t,\widehat{\tau}_t)$ be the observable decision context,
where
$\widehat{\tau}_t=(\widehat{\boldsymbol{\omega}}_t,\widehat{B}_t)$. \algname\
produces the fused task estimate
$\widetilde{\tau}_t=(\widetilde{\boldsymbol{\omega}}_t,\widetilde B_t)$, where
$\widetilde B_t$ represents the fused instantaneous admissible cost level. For
a policy conditioned on this fused context, define $\bar c_{\pi,t}
:=
\mathbb{E}_{a\sim\pi(\cdot\mid x_t,\widetilde{\tau}_t)}
[c(\mathbf{s}_t,a)]$ and $\widehat c_{\pi,t}
:=
\mathbb{E}_{a\sim\pi(\cdot\mid x_t,\widetilde{\tau}_t)}
\left[
\mathbb{E}_{p^c_\nu(\cdot\mid x_t,a)}[c]
\right]$ as the true and model-predicted one-step costs, where $p^c_\nu$ is the distributional cost model parameterized by $\nu$. The resulting perturbation terms are as follows
\begin{equation}
\delta_t^\omega
=
\left\|
\widetilde{\boldsymbol{\omega}}_t-\boldsymbol{\omega}_t
\right\|_2,
\qquad
\delta_t^B
=
\left|
\widetilde B_t-B_t
\right|,
\qquad
\epsilon_t^c(\pi)
=
\left|
\widehat c_{\pi,t}-\bar c_{\pi,t}
\right|.
\label{eq:theory-errors}
\end{equation}

\begin{assumption}
\label{ass:theory-bounded}
For all state-action pairs $(\mathbf{s},\mathbf{a})$,
$\|\mathbf{r}(\mathbf{s},\mathbf{a})\|_2\le R_2$ and
$0\le c(\mathbf{s},\mathbf{a})\le C_{\max}$. The context-dependent dual penalty is parameterized or clipped so that it is
uniformly bounded as
$0\le \lambda_\psi(\widetilde B_t,\mathbf{z}_t)\le \Lambda$.
\end{assumption}

\begin{proposition}
\label{prop:theory-fusion}
Assume that $\Pi_\Omega$ is the Euclidean projection onto the preference simplex
and that $0\le\alpha_{t,i}\le1$ for all objectives. The inferred budget is
parameterized as $B_t^{\mathrm{inf}}=\varphi(g_B(\mathbf{z}_t))$ with
$\varphi(x)=\log(1+\exp(x))$, and the exposed budget $\widehat{B}_t$ is clipped
to be nonnegative. Define
$\alpha_t^-=\min_i\{\alpha_{t,i}\}$ and
$\alpha_t^+=\max_i\{\alpha_{t,i}\}$. Let $\alpha_t^B\in[0,1]$ denote the budget
gate; in our implementation, $\alpha_t^B=\bar \alpha_t = K^{-1}\sum_{i=1}^K\alpha_{t,i}$.
Then $\widetilde{\boldsymbol{\omega}}_t\in\Omega$ and
$\widetilde B_t\ge0$. Moreover,
\begin{equation}
\delta_t^\omega
\le
(1-\alpha_t^-)
\left\|
\widehat{\boldsymbol{\omega}}_t-\boldsymbol{\omega}_t
\right\|_2
+
\alpha_t^+
\left\|
\boldsymbol{\omega}^{\mathrm{inf}}_t-\boldsymbol{\omega}_t
\right\|_2, \delta_t^B
\le
(1-\alpha_t^B)
\left|
\widehat{B}_t-B_t
\right|
+
\alpha_t^B
\left|
B^{\mathrm{inf}}_t-B_t
\right|.
\label{eq:theory-fusion-pref}
\end{equation}

\end{proposition}

Proposition~\ref{prop:theory-fusion} captures the intended arbitration behavior of \algname: uncertain beliefs are downweighted in favor of the exposed descriptor, while confident and accurate beliefs can override stale or noisy task
signals. We next connect this local fusion property to control performance, demonstrating that both preference loss and instantaneous safety violation are controlled by three errors explicitly targeted by the method: preference-fusion
error, budget-fusion error, and cost-model error. Define the true
instantaneous-penalized performance and its fused model-based counterpart as follows
\begin{equation}
\mathcal{G}_{\tau_{0:T-1}}(\pi)
:=
\mathbb{E}_{\pi,\widetilde{\tau}}
\left[
\sum_{t=0}^{T-1}
\gamma^t
\left\{
\boldsymbol{\omega}_t^\top\mathbf{r}_t
-
\lambda_\psi(\widetilde B_t,\mathbf{z}_t)
\left[
\bar c_{\pi,t}-B_t
\right]_+
\right\}
\right],
\label{eq:true-penalized-performance}
\end{equation}
\begin{equation}
\widehat{\mathcal{G}}_{\widetilde{\tau}_{0:T-1}}(\pi)
:=
\mathbb{E}_{\pi,\widetilde{\tau}}
\left[
\sum_{t=0}^{T-1}
\gamma^t
\left\{
\widetilde{\boldsymbol{\omega}}_t^\top\mathbf{r}_t
-
\lambda_\psi(\widetilde B_t,\mathbf{z}_t)
\left[
\widehat c_{\pi,t}-\widetilde B_t
\right]_+
\right\}
\right].
\label{eq:fused-penalized-performance}
\end{equation}
Here, $[x]_+=\max(x,0)$ denotes the positive-part operator. Thus,
$[\widehat c_{\pi,t}-\widetilde B_t]_+$ measures the predicted instantaneous constraint violation: it is zero when the predicted one-step cost is below the fused admissible cost level, and increases only when the predicted cost exceeds
$\widetilde B_t$.
\begin{theorem}
\label{thm:theory-robustness}
Under Assumption~\ref{ass:theory-bounded}, for any history-conditioned policy
$\pi$, the gap between the true instantaneous-penalized performance and the
fused model-based performance is bounded by
\begin{equation}
\left|
\mathcal{G}_{\tau_{0:T-1}}(\pi)
-
\widehat{\mathcal{G}}_{\widetilde{\tau}_{0:T-1}}(\pi)
\right|
\le
\sum_{t=0}^{T-1}
\gamma^t
\mathbb{E}_{\pi,\widetilde{\tau}}
\left[
R_2\delta_t^\omega
+
\Lambda
\left(
\epsilon_t^c(\pi)+\delta_t^B
\right)
\right].
\label{eq:theory-robustness-bound}
\end{equation}
Furthermore, let $\Pi$ be a comparator policy class and suppose that
$\widetilde{\pi}\in\Pi$ is $\epsilon_{\mathrm{opt}}$-optimal for the fused
model-based objective, namely $\widehat{\mathcal{G}}_{\widetilde{\tau}_{0:T-1}}(\widetilde{\pi})
\ge
\sup_{\pi\in\Pi}
\widehat{\mathcal{G}}_{\widetilde{\tau}_{0:T-1}}(\pi)
-
\epsilon_{\mathrm{opt}}$. If the perturbation terms are uniformly bounded in expectation over $\Pi$, i.e.,
$\forall$ $\pi\in\Pi$, $\forall t$, $\mathbb{E}_{\pi,\widetilde{\tau}}
\left[
\delta_t^\omega
\right]
\le
\bar{\delta}_t^\omega,
\mathbb{E}_{\pi,\widetilde{\tau}}
\left[
\delta_t^B
\right]
\le
\bar{\delta}_t^B,
\mathbb{E}_{\pi,\widetilde{\tau}}
\left[
\epsilon_t^c(\pi)
\right]
\le
\bar{\epsilon}_t^c$,
then the true suboptimality of $\widetilde{\pi}$ satisfies
\begin{equation}
\sup_{\pi\in\Pi}
\left\{
\mathcal{G}_{\tau_{0:T-1}}(\pi)
\right\}
-
\mathcal{G}_{\tau_{0:T-1}}(\widetilde{\pi})
\le
\epsilon_{\mathrm{opt}}
+
2
\sum_{t=0}^{T-1}
\gamma^t
\left[
R_2\bar{\delta}_t^\omega
+
\Lambda
\left(
\bar{\epsilon}_t^c+\bar{\delta}_t^B
\right)
\right].
\label{eq:theory-regret-bound}
\end{equation}
\end{theorem}

Theorem~\ref{thm:theory-robustness} indicates that partial task observability affects
performance only through fused-context errors. Preference mismatch drives reward
loss, while cost-model and budget-fusion errors govern instantaneous safety degradation. This matches \algname, which uses self-supervised belief inference,
uncertainty-gated fusion, and context-dependent dual penalties to reduce the observable proxies of these errors under partial task observability.

\begin{assumption}
\label{ass:selfsup-identifiability}
For a fixed latent task $\tau=(\boldsymbol{\omega},B)$, the observable history
distribution induced by
$H_t$ is informative enough to identify the
task context up to statistical error. Specifically, there exist belief decoders
$d_\omega$ and $d_B$ such that, when trained from observable prediction losses,
their fixed-task estimation errors satisfy $\mathbb{E}
\left[
\left\|
d_\omega(\mathbf{z}_t)-\boldsymbol{\omega}_t
\right\|_2
\right]
\le
A_\omega L^{-1/2}
$ and $
\mathbb{E}
\left[
\left|
d_B(\mathbf{z}_t)-B_t
\right|
\right]
\le
A_B L^{-1/2}.$
\end{assumption}
\begin{theorem} \label{thm:theory-history} Assume that the
drift targets satisfy
$\bar{\boldsymbol{\omega}}_t\in\Omega$ and
$B_t,\bar B_t\in[B_{\min},B_{\max}]$, and let
$D_B=B_{\max}-B_{\min}$. Then, under task drift,
\begin{equation}
\mathbb{E}
\left[
\left\|
\boldsymbol{\omega}^{\mathrm{inf}}_t-\boldsymbol{\omega}_t
\right\|_2
\right]
\le
A_\omega L^{-1/2}
+
\frac{L+1}{2}\sqrt{2}\rho_\omega, \mathbb{E}
\left[
\left|
B^{\mathrm{inf}}_t-B_t
\right|
\right]
\le
A_B L^{-1/2}
+
\frac{L+1}{2}D_B\rho_B .
\label{eq:theory-history-pref}
\end{equation}

Consequently, up to constants, the preference-optimal history length satisfies $L_\omega^\star
\asymp
\left(
{A_\omega}/{\rho_\omega}
\right)^{2/3}$ and  $\min_L
\mathbb{E}
\left[
\left\|
\boldsymbol{\omega}^{\mathrm{inf}}_t-\boldsymbol{\omega}_t
\right\|_2
\right]
=
\mathcal{O}
\left(
A_\omega^{2/3}\rho_\omega^{1/3}
\right),$ with the analogous budget scaling
$L_B^\star\asymp(A_B/(D_B\rho_B))^{2/3}$.
\end{theorem}

Theorem~\ref{thm:theory-history} highlights a bias-variance trade-off in history-based inference. Longer histories reduce statistical uncertainty but increase staleness bias under task drift. This motivates uncertainty-gated
fusion, which downweights inferred task estimates when the history-derived belief is unreliable.
 
\section{Experiments}
\label{sec:experiments}
\subsection{Experimental Setting}
\label{subsec:exp-setup}
\paragraph{Benchmarks.} We evaluate \algname{} on a diverse set of continuous and discrete MORL
benchmarks. The continuous dataset includes MuJoCo robotic-control tasks, including Ant, HalfCheetah, and Humanoid, and the Building energy-control task, which involves up to nine objectives \cite{li2026cola,liu2025efficient}. The discrete suite includes Minecart and
MO-LunarLander \cite{liu2025efficient}. Detailed benchmark descriptions are provided in Appendix~\ref{app:env_setup}.

\paragraph{Baselines.}
\algname{} is compared against several state-of-the-art MORL baselines, including GPI-LS \cite{gpi-ls}, C-MORL \cite{liu2025efficient}, Q-Pensieve \cite{hung2023qpensieve}, and COLA \cite{li2026cola}. For each baseline, we use the official implementation when available, tune its hyperparameters, and report the best validated setting. All methods are trained for $1$ million environment steps, with performance
reported as mean $\pm$ standard deviation over $5$ independent runs. It is noted that GPI-LS supports both discrete and continuous control, while the other baselines are designed mainly for continuous-control MORL. Additional implementation details
are given in Appendix~\ref{app:implementation_details}.

\paragraph{Metrics.}
We evaluate all methods in both static and drift settings. In the static setting,
preferences and budgets are fixed, and performance is measured by feasible-front
quality. Let $\mathcal{P}$ denote the feasible Pareto front, containing only
return vectors that satisfy the cost threshold.
\begin{itemize}
    \item \textbf{Feasible Hypervolume (FHV)} \cite{li2026cola} measures the dominated volume
    covered by feasible Pareto solutions with respect to a reference point
    $\mathbf{r}^{\mathrm{ref}}\in\mathbb{R}^{K}$:
    $
    \mathrm{FHV}
    =
    \mathrm{HV}(\mathcal{P};\mathbf{r}^{\mathrm{ref}})
    =
    \lambda_K\left(
    \bigcup_{\mathbf{R}\in\mathcal{P}}
    \prod_{k=1}^{K}[r^{\mathrm{ref}}_k,R_k]
    \right),
    $
    where $\lambda_K(\cdot)$ denotes the $K$-dimensional Lebesgue measure.

    \item \textbf{Feasible Utility (FUT)} \cite{li2026cola} evaluates the best feasible scalarized
    return under evaluation preferences
    $\Omega_{\mathrm{eval}}=\{\boldsymbol{\omega}_j\}_{j=1}^{M}\subset\Omega$:
    $
    \mathrm{FUT}
    =
    \frac{1}{M}\sum_{j=1}^{M}
    \max_{\mathbf{R}\in\mathcal{P}}
    \boldsymbol{\omega}_j^\top \mathbf{R}.
    $
\end{itemize}
In the drift setting, preferences and budgets vary during evaluation, so we
evaluate online performance against the current task variables. For episode
$e\in\{1,\ldots,E\}$ and timestep $t\in\{0,\ldots,T_e-1\}$, let
$\boldsymbol{\omega}_{e,t}\in\Omega$ denote the current hidden preference,
$B_{e,t}$ the current budget, $\mathbf{r}_{e,t}$ the reward vector, and
$c_{e,t}$ the realized cost.
\begin{itemize}
    \item \textbf{Scalarized Return (SR)} \cite{peng2023multi} measures average online scalarized
    performance:
    $
    \mathrm{SR}
    =
    \frac{1}{E}
    \sum_{e=1}^{E}
    \frac{1}{T_e}
    \sum_{t=0}^{T_e-1}
    \boldsymbol{\omega}_{e,t}^{\top}\mathbf{r}_{e,t}.
    $

    \item \textbf{Post-Change Utility (PCU)} \cite{luo2022adapt} measures utility during the
    adaptation window after preference or budget changes:
    $
    \mathrm{PCU}
    =
    \frac{1}{E}
    \sum_{e=1}^{E}
    \frac{1}{\max(|\mathcal{A}_e|,1)}
    \sum_{t\in\mathcal{A}_e}
    \boldsymbol{\omega}_{e,t}^{\top}\mathbf{r}_{e,t},
    $
    where $\mathcal{A}_e$ denotes the post-change adaptation window.
\end{itemize}
Additionally, we utilize \textbf{Mean Cost (MC)} and \textbf{Violation Rate (VR)}
to monitor constraint satisfaction.

\subsection{Experimental Results}
\begin{table}[t]
\centering
\caption{
Reference-threshold feasible-front evaluation of \algname{} against MORL baselines}
\vspace{0.5em}
\label{tab:main_feasible_results}
\setlength{\tabcolsep}{6pt}
\begin{adjustbox}{width=\textwidth}
\begin{tabular}{llccccc}
\toprule
\textbf{Dataset} & \textbf{Metric}
& \textbf{GPI-LS}
& \textbf{C-MORL}
& \textbf{Q-Pensieve}
& \textbf{COLA}
& \textbf{\algname{} (ours)} \\
\midrule

\multirow{3}{*}{MO-Ant-2d}
& FHV ($\times 10^{6}$)
& $2.61\pm0.23$ & $0.25\pm0.01$ & $5.56\pm0.34$ & $6.37\pm0.89$ & $\mathbf{7.71\pm0.10}$ \\
& FUT ($\times 10^{3}$)
& $1.99\pm0.05$ & $0.50\pm0.01$ & $2.21\pm0.07$ & $2.35\pm0.14$ & $\mathbf{2.57\pm0.02}$ \\
& MC ($\times 10^{-3}$)
& $1.91\pm0.10$ & $\mathbf{0.07\pm0.05}$ & $2.66\pm0.17$ & $2.60\pm0.81$ & $1.39\pm0.06$ \\
\midrule

\multirow{3}{*}{MO-HalfCheetah-2d}
& FHV ($\times 10^{6}$)
& $3.16\pm0.05$ & $1.25\pm0.00$ & $5.61\pm0.04$ & $5.40\pm0.12$ & $\mathbf{5.74\pm0.06}$ \\
& FUT ($\times 10^{3}$)
& $2.25\pm0.02$ & $1.50\pm0.00$ & $2.28\pm0.04$ & $2.24\pm0.04$ & $\mathbf{2.35\pm0.00}$ \\
& MC ($\times 10^{-3}$)
& $0.59\pm0.04$ & $\mathbf{0.01\pm0.00}$ & $1.49\pm0.26$ & $1.99\pm0.18$ & $0.78\pm0.01$ \\
\midrule

\multirow{3}{*}{MO-Hopper-2d}
& FHV ($\times 10^{7}$)
& $0.00\pm0.00$ & $0.13\pm0.11$ & $0.24\pm0.01$ & $0.00\pm0.00$ & $\mathbf{1.20\pm0.16}$ \\
& FUT ($\times 10^{3}$)
& $0.00\pm0.00$ & $0.97\pm0.74$ & $1.48\pm0.02$ & $0.00\pm0.00$ & $\mathbf{3.37\pm0.21}$ \\
& MC ($\times 10^{-2}$)
& $1.85\pm0.12$ & $\mathbf{0.19\pm0.16}$ & $0.46\pm0.04$ & $1.57\pm0.10$ & $0.72\pm0.08$ \\
\midrule

\multirow{3}{*}{MO-Humanoid}
& FHV ($\times 10^{7}$)
& $0.00\pm0.00$ & $0.01\pm0.00$ & $1.06\pm0.51$ & $0.71\pm0.27$ & $\mathbf{2.11\pm0.09}$ \\
& FUT ($\times 10^{3}$)
& $0.00\pm0.00$ & $0.30\pm0.05$ & $3.14\pm0.79$ & $2.74\pm0.48$ & $\mathbf{4.69\pm0.07}$ \\
& MC ($\times 10^{-3}$)
& $1.60\pm0.09$ & $\mathbf{0.04\pm0.00}$ & $0.59\pm0.02$ & $0.77\pm0.04$ & $0.07\pm0.02$ \\
\midrule

\multirow{3}{*}{MO-Ant-3d}
& FHV ($\times 10^{10}$)
& $0.31\pm0.03$ & $0.00\pm0.00$ & $0.26\pm0.06$ & $0.91\pm0.16$ & $\mathbf{1.01\pm0.03}$ \\
& FUT ($\times 10^{3}$)
& $1.55\pm0.01$ & $0.69\pm0.01$ & $1.29\pm0.08$ & $2.00\pm0.08$ & $\mathbf{2.05\pm0.01}$ \\
& MC ($\times 10^{-3}$)
& $3.22\pm0.27$ & $\mathbf{0.06\pm0.01}$ & $3.62\pm0.10$ & $3.47\pm1.08$ & $1.44\pm0.08$ \\
\midrule

\multirow{3}{*}{MO-Ant-4d}
& FHV ($\times 10^{14}$)
& $3.52\pm0.51$ & $0.17\pm0.00$ & $2.35\pm1.01$ & $3.25\pm0.78$ & $\mathbf{4.78\pm0.15}$ \\
& FUT ($\times 10^{3}$)
& $4.32\pm0.33$ & $2.03\pm0.01$ & $3.79\pm0.45$ & $4.21\pm0.27$ & $\mathbf{4.65\pm0.04}$ \\
& MC ($\times 10^{-3}$)
& $1.56\pm0.23$ & $\mathbf{0.08\pm0.05}$ & $2.73\pm0.21$ & $3.22\pm1.04$ & $1.64\pm0.14$ \\
\midrule

\multirow{3}{*}{MO-Ant-5d}
& FHV ($\times 10^{16}$)
& $0.21\pm0.05$ & $0.11\pm0.02$ & $2.53\pm0.54$ & $2.08\pm0.51$ & $\mathbf{3.45\pm0.04}$ \\
& FUT ($\times 10^{3}$)
& $1.87\pm0.09$ & $1.61\pm0.00$ & $1.91\pm0.09$ & $1.86\pm0.10$ & $\mathbf{2.04\pm0.01}$ \\
& MC ($\times 10^{-3}$)
& $1.09\pm0.13$ & $\mathbf{0.07\pm0.02}$ & $1.19\pm0.14$ & $1.11\pm0.45$ & $0.42\pm0.02$ \\
\midrule

\multirow{3}{*}{MO-HalfCheetah-5d}
& FHV ($\times 10^{17}$)
& $0.22\pm0.00$ & $0.01\pm0.02$ & $0.30\pm0.19$ & $0.33\pm0.05$ & $\mathbf{1.22\pm0.07}$ \\
& FUT ($\times 10^{3}$)
& $2.29\pm0.05$ & $1.57\pm0.04$ & $2.12\pm0.26$ & $2.59\pm0.08$ & $\mathbf{2.93\pm0.07}$ \\
& MC ($\times 10^{-3}$)
& $2.93\pm0.25$ & $2.46\pm0.68$ & $2.96\pm0.56$ & $3.81\pm0.94$ & $\mathbf{1.66\pm0.03}$ \\
\midrule

\multirow{3}{*}{Building-9d}
& FHV ($\times 10^{31}$)
& $4.52 \pm 0.11$ & $5.02\pm0.57$ & $6.32\pm0.57$ & $5.84\pm0.27$ & $\mathbf{6.98\pm0.15}$ \\
& FUT ($\times 10^{3}$)
& $3.33 \pm 0.02$ & $3.36\pm0.03$ & $3.42\pm0.03$ & $3.39\pm0.02$ & $\mathbf{3.45\pm0.01}$ \\
& MC ($\times 10^{-3}$)
& $2.96 \pm 0.21$ & $\mathbf{0.09\pm0.12}$ & $6.29\pm0.30$ & $5.44\pm0.17$ & $0.54\pm0.28$ \\

\bottomrule
\end{tabular}
\end{adjustbox}
\end{table}

\begin{figure*}[t]
    \centering

    \begin{minipage}{0.24\linewidth}
        \centering
        \includegraphics[width=\linewidth]{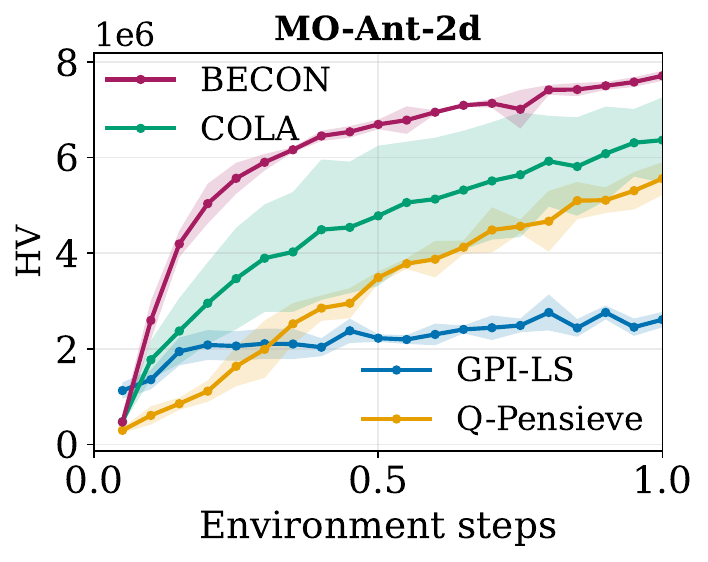}
    \end{minipage}
    \hfill
    \begin{minipage}{0.24\linewidth}
        \centering
        \includegraphics[width=\linewidth]{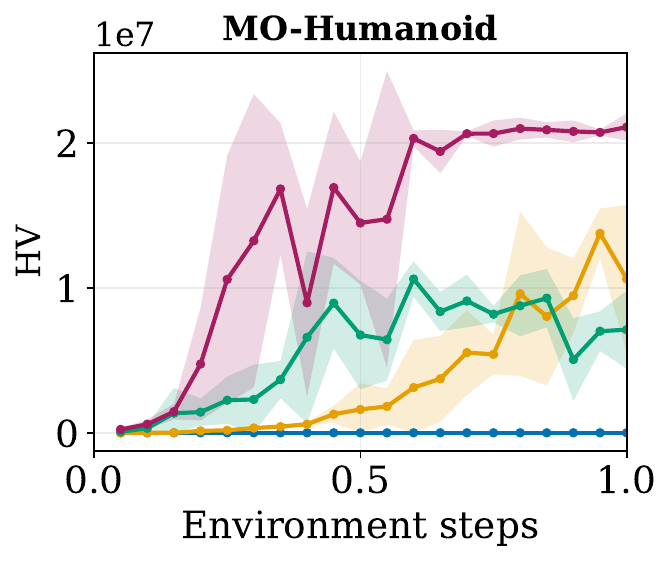}
    \end{minipage}
    \hfill
    \begin{minipage}{0.24\linewidth}
        \centering
        \includegraphics[width=\linewidth]{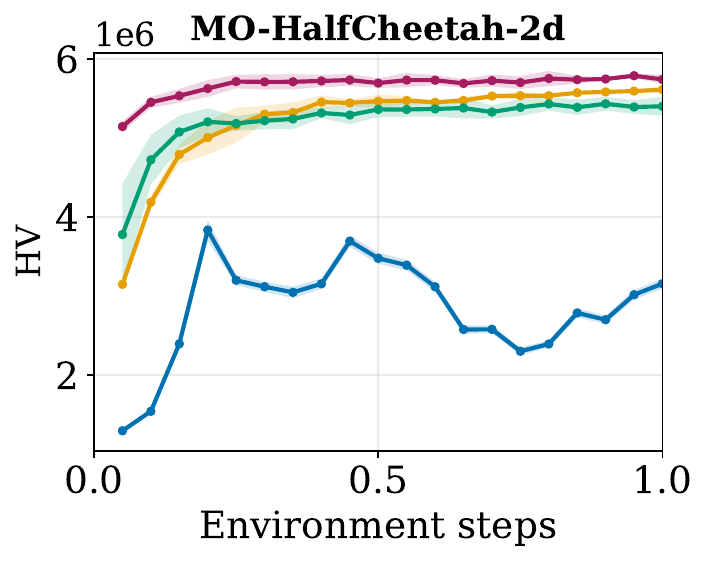}
    \end{minipage}
    \hfill
    \begin{minipage}{0.24\linewidth}
        \centering
        \includegraphics[width=\linewidth]{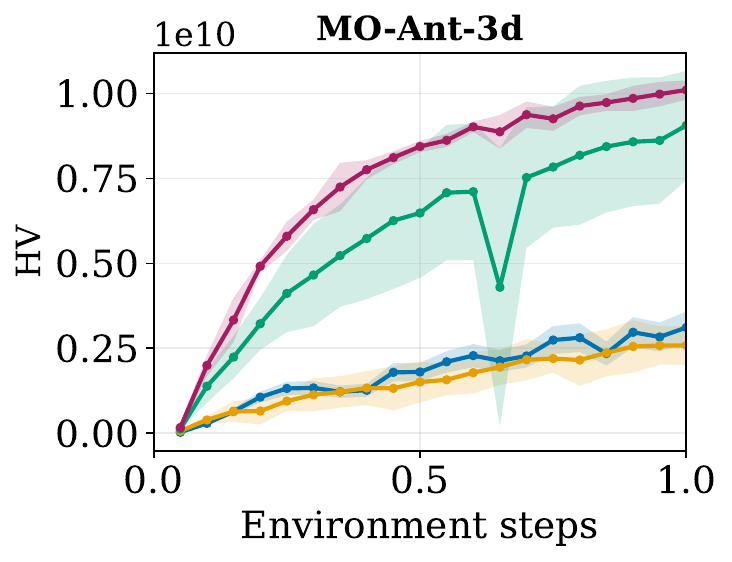}
    \end{minipage}

    \vspace{0.5em}

    \begin{minipage}{0.24\linewidth}
        \centering
        \includegraphics[width=\linewidth]{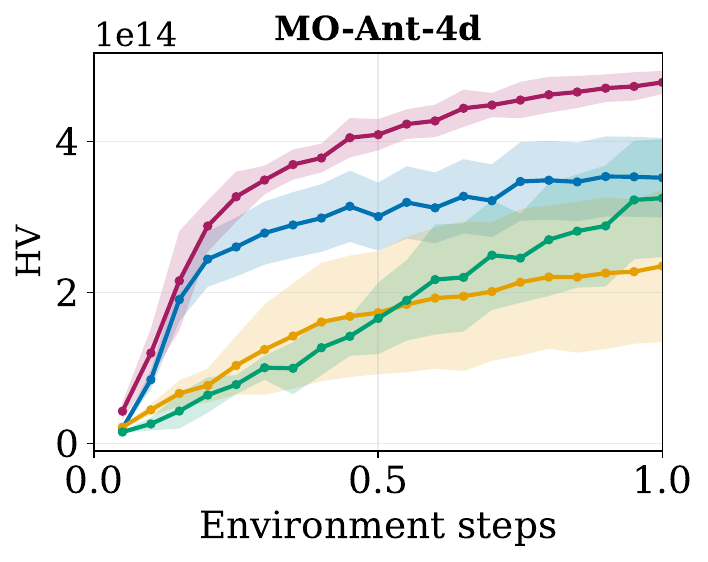}
    \end{minipage}
    \hfill
    \begin{minipage}{0.24\linewidth}
        \centering
        \includegraphics[width=\linewidth]{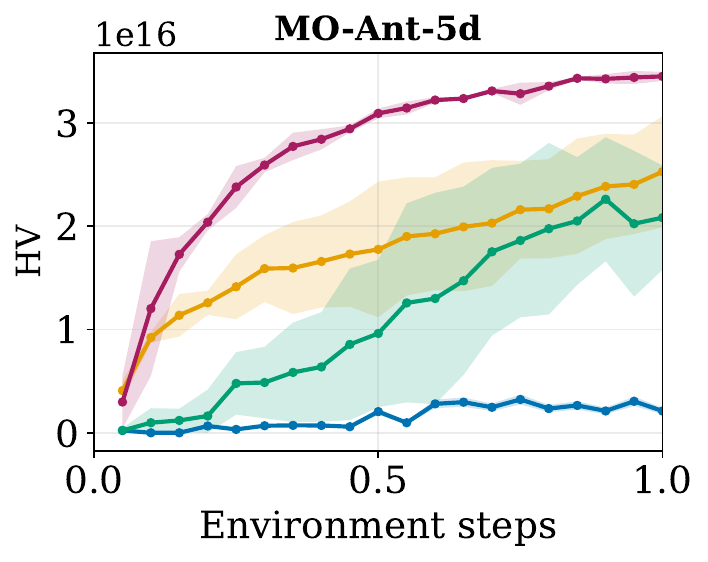}
    \end{minipage}
    \hfill
    \begin{minipage}{0.24\linewidth}
        \centering
        \includegraphics[width=\linewidth]{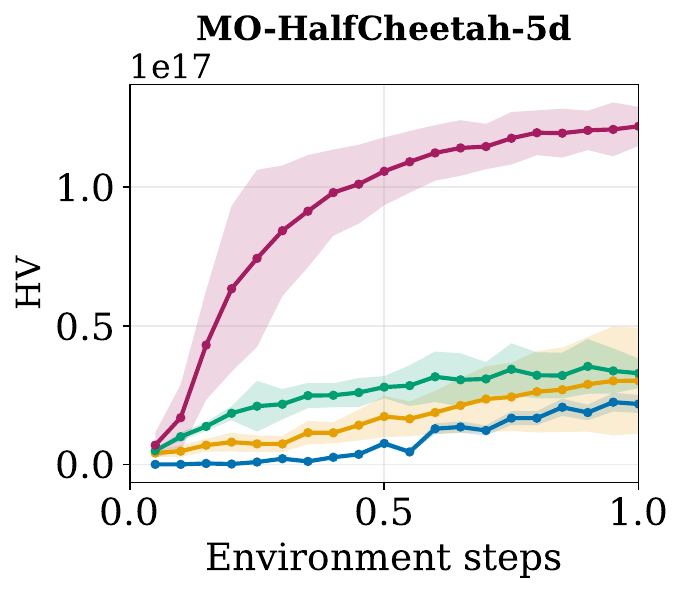}
    \end{minipage}
    \hfill
    \begin{minipage}{0.24\linewidth}
        \centering
        \includegraphics[width=\linewidth]{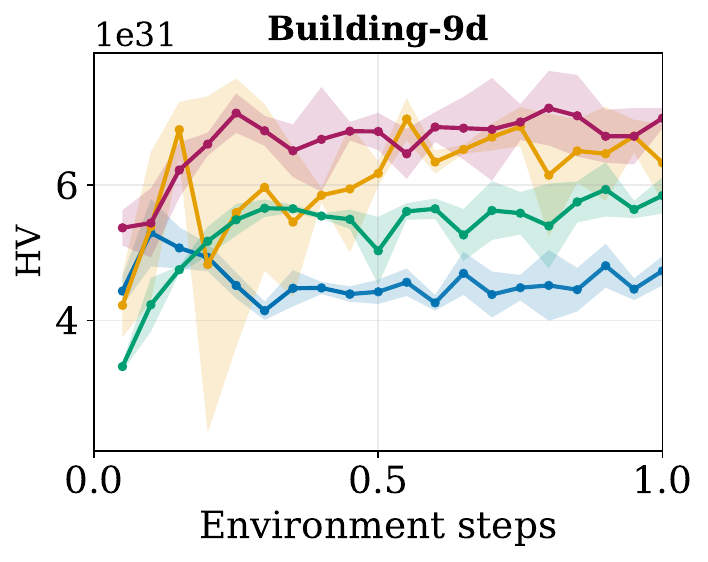}
    \end{minipage}

\vspace{-0.3cm}
\caption{Feasible hypervolume convergence of algorithms (environment steps in millions).}
    \label{fig:hv_convergence_all}
\end{figure*}

Table~\ref{tab:main_feasible_results} demonstrates reference-threshold
feasible-front performance under static preferences and budgets. It is noted that if no solution satisfies the cost threshold, the feasible front is empty and both
FHV and FUT are set to zero.
\algname{}
achieves the best FHV and FUT on all benchmark tasks, indicating that the
proposed belief-conditioned constraint handling improves both feasible-front
coverage and preference-aligned utility. Compared with the second-best method,
\algname{} improves FHV by $98.4\%$ on average and FUT by $24.5\%$ on average
across all tasks. The gains are especially pronounced on challenging tasks such as Building-9d,
MO-Humanoid, MO-Ant-5d, and MO-HalfCheetah-5d, where accurately identifying
feasible high-return solutions becomes substantially harder. Although C-MORL often obtains the lowest mean cost, this behavior reflects
conservative constraint satisfaction rather than superior feasible-front
optimization. In constrained MORL, MC is used to monitor whether the learned
policy respects the cost threshold, while the primary goal is to maximize
feasible return quality. \algname{} maintains low realized costs while
consistently achieving substantially higher FHV and FUT, highlighting a better balance between constraint satisfaction and multi-objective performance. Additional constraint-tightness results in Appendix~\ref{app:budget_sensitivity}
illustrate that \algname~preserves a nonzero FHV under stringent admissible
cost levels where COLA often becomes infeasible. Table~\ref{tab:discrete-feasible-results} further confirms this trend on
discrete benchmarks. \algname{} obtains the best FHV and FUT on both MO-LunarLander-4d and Minecart, while keeping MC well below the feasibility
threshold.
Figure~\ref{fig:hv_convergence_all} further shows the training dynamics in terms
of feasible hypervolume. Across all environments, \algname{} reaches higher FHV
with more stable convergence than the baselines. The smaller uncertainty bands
indicate lower variance across seeds, while the consistently higher curves show
better sample efficiency and more reliable discovery of feasible Pareto solutions.

\begin{table}
\vspace{0.5em}
\centering
\caption{Feasible-front performance on discrete MORL benchmarks}
\label{tab:discrete-feasible-results}
\scriptsize
\setlength{\tabcolsep}{3.9pt}
\begin{tabular}{llccc}
\toprule
\textbf{Dataset} & \textbf{Method} & \textbf{FHV} $\uparrow$ & \textbf{FUT} $\uparrow$ & \textbf{MC} \\
\midrule
\multirow{4}{*}{MO-LunarLander-4d}
& \algname{} & $\mathbf{3.69{\times}10^{10}}$ & $\mathbf{71.58}$ & 0.00102 \\
& COLA & $3.44{\times}10^{10}$ & 68.98 & 0.00379 \\
& Q-Pensieve & $3.36{\times}10^{10}$ & 67.48 & 0.00363 \\
& GPI-LS & $1.96{\times}10^{10}$ & 66.72 & 0.00324 \\
\midrule
\multirow{4}{*}{Minecart}
& \algname{} & $\mathbf{189.70}$ & $\mathbf{-6.07}$ & 0.00001 \\
& COLA & 180.00 & -6.67 & 0.00695 \\
& Q-Pensieve & 179.90 & -6.70 & 0.00404 \\
& GPI-LS & 152.90 & -8.70 & 0.00313 \\
\bottomrule
\end{tabular}
\end{table}

Table~\ref{tab:drift-story-results} evaluates online adaptation under drifting
preferences and budgets. \algname{} achieves the best SR and PCU on all tasks,
demonstrating stronger online utility and faster post-change recovery than the
baselines. \algname{} also achieves zero violation rate and the lowest MC across all
benchmarks, suggesting that its adaptation gains are obtained under stringent
constraint compliance rather than relaxed feasibility. These results demonstrate that
belief-conditioned task inference and context-dependent dual control enable
robust adaptation to task drift.

\begin{table*}[t]
\centering
\caption{Task-drift adaptation results}
\label{tab:drift-story-results}
\tiny
\setlength{\tabcolsep}{4pt}
\renewcommand{\arraystretch}{1.12}
\begin{tabular}[t]{llcccc@{\hspace{0.35cm}}|@{\hspace{0.35cm}}llcccc}
\hline
\textbf{Dataset} & \textbf{Method} & \textbf{SR} $\uparrow$ & \textbf{PCU} $\uparrow$ & \textbf{VR} $\downarrow$ & \textbf{MC} $\downarrow$
&
\textbf{Dataset} & \textbf{Method} & \textbf{SR} $\uparrow$ & \textbf{PCU} $\uparrow$ & \textbf{VR} $\downarrow$ & \textbf{MC} $\downarrow$ \\
\hline

\multirow{4}{*}{MO-Ant-2d}
& \algname{} & \best{4.68} & \best{4.76} & 0 & \best{0.00142}
& \multirow{4}{*}{MO-Humanoid}
& \algname{} & \best{4.45} & \best{4.49} & 0 & \best{0.00015} \\
& COLA & 4.32 & 4.38 & 0 & 0.00217
& & COLA & 2.23 & 2.19 & 0 & 0.00069 \\
& Q-Pensieve & 3.56 & 3.60 & 0 & 0.00279
& & Q-Pensieve & 3.37 & 3.33 & 0 & 0.00037 \\
& GPI-LS & 3.49 & 3.63 & 0 & 0.00185
& & GPI-LS & -3.11 & 0.00 & 0 & 0.00160 \\
\hline

\multirow{4}{*}{MO-Ant-3d}
& \algname{} & \best{3.64} & \best{3.71} & 0 & \best{0.00168}
& \multirow{4}{*}{MO-HalfCheetah-2d}
& \algname{} & \best{4.58} & \best{4.58} & 0 & \best{0.00080} \\
& COLA & 3.17 & 3.25 & 0 & 0.00506
& & COLA & 4.31 & 4.32 & 0 & 0.00159 \\
& Q-Pensieve & 1.54 & 1.59 & 0.0008 & 0.00424
& & Q-Pensieve & 4.41 & 4.40 & 0 & 0.00117 \\
& GPI-LS & 2.93 & 2.98 & 0 & 0.00346
& & GPI-LS & 4.11 & 4.08 & 0 & 0.00167 \\
\hline

\multirow{4}{*}{MO-Ant-4d}
& \algname{} & \best{9.33} & \best{9.44} & 0 & \best{0.00151}
& \multirow{4}{*}{MO-HalfCheetah-5d}
& \algname{} & \best{5.63} & \best{5.81} & 0 & \best{0.00205} \\
& COLA & 6.71 & 6.86 & 0 & 0.00368
& & COLA & 5.03 & 3.51 & 0.0023 & 0.00591 \\
& Q-Pensieve & 7.31 & 7.51 & 0 & 0.00297
& & Q-Pensieve & 3.28 & 1.38 & 0 & 0.00474 \\
& GPI-LS & 8.57 & 8.76 & 0 & 0.00158
& & GPI-LS & 4.06 & 4.11 & 0.0024 & 0.00436 \\
\hline

\multirow{4}{*}{MO-Ant-5d}
& \algname{} & \best{3.75} & \best{3.73} & 0 & \best{0.00082}
& \multirow{4}{*}{Building-9d}
& \algname{} & \best{7.08} & \best{7.09} & 0 & \best{0.00017} \\
& COLA & 2.81 & 2.75 & 0 & 0.00124
& & COLA & 7.03 & 7.05 & 0 & 0.00566 \\
& Q-Pensieve & 3.20 & 3.16 & 0 & 0.00169
& & Q-Pensieve & 6.86 & 6.86 & 0 & 0.00675 \\
& GPI-LS & 2.48 & 2.41 & 0 & 0.00194
& & GPI-LS & 6.25 & 6.21 & 0 & 0.00635 \\
\hline
\end{tabular}
\end{table*}

Additional analyses in Appendix~\ref{app:additional_experiments} examine the
robustness of \algname{} and its main design choices. In particular,
Appendix~\ref{app:component_ablation} validates the contribution of belief
inference and feasibility-aware control, Appendix~\ref{app:history_encoder_ablation}
compares temporal encoders, and Appendix~\ref{app:history_length_ablation}
studies the effect of history length $L$. Appendix~\ref{app:intent_mix_ablation}
and Appendix~\ref{app:preference_drift_ablation} analyze sensitivity to the
fusion weight $\alpha_{\max}$ and preference drift rate $\rho_\omega$,
respectively. Finally, Appendix~\ref{app:dual_ablation} shows that the full
context-dependent dual $\lambda_\psi(\widetilde B_t,\mathbf{z}_t)$ provides the
most consistent balance between feasible-front quality and online adaptation.

\section{Limitations}
\label{sec:limitations}
BECON focuses on stage-wise admissible cost levels rather than cumulative or
remaining-budget constraints. This formulation is useful when safety tolerance is
context-dependent and must be enforced locally at each decision step, such as
when acceptable energy use, risk exposure, or operational cost changes with user
preferences, environmental conditions, or recent interaction history.
By treating $B_t$ as an instantaneous admissible cost level, the agent can adapt
constraint enforcement to the current task context instead of relying on a fixed
global budget. However, this does not directly capture long-horizon
resource depletion or trade-offs with future remaining budgets.
Extending the framework to cumulative constraints would require augmenting the
state with remaining budget dynamics and developing belief updates over both
task context and residual feasibility. BECON assumes that latent preferences and budgets are
sufficiently identifiable from finite interaction histories, which may not hold
under uninformative or stale observations. Our experiments isolate
task-specification drift under fixed environment dynamics, leaving joint
task-environment drift and real-world deployment unexplored.

\section{Conclusion}
\label{sec:conclusion}
This paper studies constrained preference-conditioned MORL with drifting preferences and
budgets as a partially observed task-specification problem. We introduced
\algname{}, a belief-conditioned framework that infers latent task context from
recent histories, uncertainty-gates inferred and exposed task signals, and
adapts constraint enforcement through a context-dependent dual variable. Our
analysis decomposes performance loss into preference-fusion, budget-fusion, and
cost-model errors, while characterizing the history-length trade-off under drift. \algname{}
consistently improves feasible-front quality, online adaptation, and constraint
satisfaction over strong MORL baselines across various MORL benchmarks. These results suggest that preferences and budgets should be treated as uncertain task observations rather than fixed
ground-truth commands.

\section*{Acknowledgments and Disclosures}

\noindent\textbf{Funding.}
This research was funded by Hanoi University of Science
and Technology under project code T2025-TN-005.

\noindent\textbf{Competing Interests.}
The authors declare no competing financial interests
related to this work.
\newpage
\bibliographystyle{unsrtnat}
\bibliography{references}

\newpage
\appendix

\section{Additional Related Work}
\label{app:related}

\paragraph{MORL and preference-conditioned control.}
MORL studies sequential decision making
with vector-valued returns, where policies must negotiate competing objectives
rather than optimize a single scalar reward. Linear scalarization remains a
dominant paradigm because it reduces MORL to standard RL under a preference
vector, which can recover broad classes of Pareto-optimal solutions under
suitable structural assumptions \cite{lu2023multiobjective}. However, learning
a separate policy for each preference is inefficient and poorly scalable.
Preference-conditioned methods address this issue by training universal or
meta-policies that take preferences as inputs and generalize across the
preference simplex \cite{basaklar2023pdmorl}. Recent work further improves
Pareto-front coverage through constrained Pareto discovery, latent knowledge
sharing, and preference-controllable optimization \cite{liu2025efficient,
pmlr-v267-yang25ax}. Distributional MORL extends this view by optimizing over
return distributions rather than only expected returns, thereby capturing richer
risk-sensitive trade-offs \cite{NEURIPS2023_32285dd1}. A complementary line of
work considers changing objectives, including continual MORL and latent
preference inference under non-stationary priorities \cite{li2024continual,
cao2026learning}. Despite these advances, most preference-conditioned methods
still assume that the operative preference is either directly observed, fixed
within the decision context, or inferable without explicitly modeling feasibility.

\paragraph{Constrained and safe RL.}
Constrained RL augments reward maximization with safety,
resource, or budget constraints. Budgeted MDPs formalize this trade-off through
a cost signal and an adjustable threshold, enabling policies to condition their
behavior on remaining risk tolerance \cite{carrara2019budgeted}. In CMDPs,
primal-dual and Lagrangian approaches are widely used because they convert
constraint satisfaction into a penalty-weighted optimization problem, but their
performance can be sensitive to cost-estimation bias and multiplier dynamics
\cite{ICLR2024_2f8b5654}. Recent theoretical work has refined constrained RL
through reward-free exploration, approachability, and adaptive resource
resolving, yielding sharper guarantees in tabular and structured settings
\cite{pmlr-v162-miryoosefi22a,jiang2024achieving}. In practical safe RL,
off-policy and offline methods address sample efficiency, OOD actions, and
safety during deployment by using conservative cost estimation, constrained
exploration, or minimax formulations \cite{li2026offpolicy,chemingui2026online}.
Adaptive-threshold methods further condition policies or policy selection on
varying safety limits, but they usually treat the constraint threshold as an
explicit input rather than a partially observed quantity \cite{guo2025constraint,
chemingui2025constraint}. When constraints are combined with multiple
objectives, recent algorithms cast objective conflicts and safety requirements
as a unified constrained optimization problem \cite{ICLR2025_59ddfff7}. Yet the
safety trade-off is often governed by a global or weakly contextualized
multiplier, which is restrictive when both preferences and budgets evolve along
a trajectory.

\paragraph{Latent context inference and non-stationary RL.}
Latent representation learning has become a promising technique for improving
generalization and sample efficiency in RL. Self-supervised temporal prediction
encourages agents to learn representations that preserve control-relevant
dynamics across future states \cite{schwarzer2021dataefficient}. Bisimulation
metrics provide another principled route by learning latent spaces that retain
behaviorally relevant information while discarding nuisance variation
\cite{zhang2021learning}. For non-stationary environments, recent methods learn
representations that expose the causal origin of distributional shifts or
regularize policies locally against catastrophic forgetting
\cite{pmlr-v235-zhang24ah,hamadanian2025online}. In-context RL similarly uses
recent trajectories to infer the active task at test time without gradient
updates, although many offline formulations remain brittle under distribution
shift in the context-collection policy \cite{pmlr-v267-ma25x}. These approaches
show that history can encode hidden task structure, but they do not directly
integrate latent context inference with constrained MORL, where both user intent
and feasible safety budget may be uncertain.

\section{Training Procedure}
\label{app:training-procedure}

Algorithm~\ref{alg:becon-compact} summarizes the training procedure of
\algname{}.

\begin{algorithm}[!ht]
\caption{\algname{} Training Procedure}
\label{alg:becon-compact}
\begin{algorithmic}[1]
\Require Replay buffer $\mathcal{D}$, history length $L$, gate scale $\alpha_{\max}$
\State Initialize encoders, task decoders, actor, twin critics, cost model, dual network, and targets
\For{each environment step $t$}
    \State Encode $\mathbf{e}_t=f_{\mathrm{enc}}(\mathbf{s}_t)$ and
    $\mathbf{z}_t\sim q_\phi(\mathbf{z}_t\mid H_t)$
    \State Decode $(\boldsymbol{\omega}^{\mathrm{inf}}_t,B^{\mathrm{inf}}_t)$ from $\mathbf{z}_t$
    \State Compute uncertainty gates $\boldsymbol{\alpha}_t$ and fuse
    $(\widehat{\boldsymbol{\omega}}_t,\widehat B_t)$ with
    $(\boldsymbol{\omega}^{\mathrm{inf}}_t,B^{\mathrm{inf}}_t)$ to obtain
    $(\widetilde{\boldsymbol{\omega}}_t,\widetilde B_t)$
    \State Act with
    $\pi_\theta(\cdot\mid\mathbf{e}_t,\mathbf{z}_t,
    \widetilde{\boldsymbol{\omega}}_t,\widetilde B_t)$ and store the transition in $\mathcal{D}$
    \State Sample mini-batches and update
    $\mathcal{L}_{\mathrm{critic}}$,
    $\mathcal{L}_{\mathrm{belief}}$,
    $\mathcal{L}_{\mathrm{dyn}}$,
    $\mathcal{L}_{\mathrm{cost}}$,
    $\mathcal{L}_{\mathrm{actor}}$, and
    $\mathcal{L}_{\lambda}$
    \State Soft-update target networks
\EndFor
\end{algorithmic}
\end{algorithm}

\section{Additional Theory and Proofs}
\label{app:theory}
This appendix provides assumptions, auxiliary results, and proofs for
Section~\ref{sec:theory}. 
\subsection{Belief Sufficiency under Task Drift}
\label{app:belief-sufficiency}

The belief module in \algname\ can be viewed as an amortized, finite-dimensional
approximation to belief-state filtering over the hidden task context.

\begin{proposition}
\label{prop:app-belief-sufficiency} Let the hidden augmented state be
$m_t=(\mathbf{s}_t,\tau_t)$, where
$\tau_t=(\boldsymbol{\omega}_t,B_t)$. Suppose the environment transition is
fixed and Markovian,
$\mathbf{s}_{t+1}\sim P(\cdot\mid \mathbf{s}_t,\mathbf{a}_t)$, the task evolves
according to a Markov drift kernel
$K_\tau(\tau_{t+1}\mid\tau_t)$, and the exposed descriptor is generated from an
observation kernel
$O(\widehat{\tau}_t\mid\tau_t)$. Then the posterior belief $b_t(m)
:=
\mathbb{P}
\left(
m_t=m
\mid
o_{0:t},\mathbf{a}_{0:t-1}
\right),$ with $ o_t=(\mathbf{s}_t,\widehat{\tau}_t)$, is a sufficient statistic for optimal control. In particular, there exists an
optimal policy $\pi^\star(\mathbf{a}_t\mid b_t)$.
\end{proposition}

\begin{proof}
Given $m_t=(\mathbf{s}_t,\tau_t)$ and action $\mathbf{a}_t$, the next augmented
state distribution factors as
\begin{equation}
\mathbb{P}(m_{t+1}\mid m_t,\mathbf{a}_t)
=
P(\mathbf{s}_{t+1}\mid\mathbf{s}_t,\mathbf{a}_t)
K_\tau(\tau_{t+1}\mid\tau_t).
\end{equation}
The next observation is generated from the realized environment state together
with the task-observation kernel. Hence, once $b_t$ and $\mathbf{a}_t$ are given,
the full history provides no additional information about future hidden states,
observations, rewards, or costs. The Bayesian update maps
$(b_t,\mathbf{a}_t,o_{t+1})$ to $b_{t+1}$, so the belief process is Markov.
Dynamic programming on this belief MDP therefore yields an optimal policy that
depends on past interactions only through $b_t$.
\end{proof}

Proposition~\ref{prop:app-belief-sufficiency} justifies the use of the learned
variational belief $q_\phi(\mathbf{z}_t\mid H_t)$. Since the exact belief $b_t$
is generally intractable, \algname\ uses $\mathbf{z}_t$ as an amortized,
finite-dimensional summary of the recent interaction history.

\subsection{Proof of Proposition~\ref{prop:theory-fusion}}
\label{app:proof-fusion}

\begin{proof}
Let $\mathbf{y}_t
=
(1-\boldsymbol{\alpha}_t)\odot
\widehat{\boldsymbol{\omega}}_t
+
\boldsymbol{\alpha}_t\odot
\boldsymbol{\omega}^{\mathrm{inf}}_t$. By construction,
$\widetilde{\boldsymbol{\omega}}_t=\Pi_\Omega(\mathbf{y}_t)$, hence
$\widetilde{\boldsymbol{\omega}}_t\in\Omega$. Since
$\boldsymbol{\omega}_t\in\Omega$, we have
$\Pi_\Omega(\boldsymbol{\omega}_t)=\boldsymbol{\omega}_t$. Euclidean projection
onto a closed convex set is non-expansive, so
\begin{align}
\left\|
\widetilde{\boldsymbol{\omega}}_t-\boldsymbol{\omega}_t
\right\|_2
=
\left\|
\Pi_\Omega(\mathbf{y}_t)
-
\Pi_\Omega(\boldsymbol{\omega}_t)
\right\|_2  \le
\left\|
\mathbf{y}_t-\boldsymbol{\omega}_t
\right\|_2 .
\end{align}
For each coordinate, $y_{t,i}-\omega_{t,i}
=
(1-\alpha_{t,i})
(\widehat{\omega}_{t,i}-\omega_{t,i})
+
\alpha_{t,i}
(\omega^{\mathrm{inf}}_{t,i}-\omega_{t,i})$. Therefore,
\begin{align}
\left\|
\mathbf{y}_t-\boldsymbol{\omega}_t
\right\|_2
&\le
\left\|
(1-\boldsymbol{\alpha}_t)\odot
(
\widehat{\boldsymbol{\omega}}_t-\boldsymbol{\omega}_t
)
\right\|_2
+
\left\|
\boldsymbol{\alpha}_t\odot
(
\boldsymbol{\omega}^{\mathrm{inf}}_t-\boldsymbol{\omega}_t
)
\right\|_2
\nonumber\\
&\le
(1-\alpha_t^-)
\left\|
\widehat{\boldsymbol{\omega}}_t-\boldsymbol{\omega}_t
\right\|_2
+
\alpha_t^+
\left\|
\boldsymbol{\omega}^{\mathrm{inf}}_t-\boldsymbol{\omega}_t
\right\|_2 .
\end{align}

For the budget, $\widetilde B_t
=
(1-\alpha_t^B)\widehat{B}_t+\alpha_t^B B_t^{\mathrm{inf}}$.
Thus,
\begin{align}
\left|
\widetilde B_t-B_t
\right|
&=
\left|
(1-\alpha_t^B)(\widehat{B}_t-B_t)
+
\alpha_t^B(B_t^{\mathrm{inf}}-B_t)
\right|
\nonumber\\
&\le
(1-\alpha_t^B)
\left|
\widehat{B}_t-B_t
\right|
+
\alpha_t^B
\left|
B_t^{\mathrm{inf}}-B_t
\right|.
\end{align}
If $\widehat{B}_t\ge0$ and $B_t^{\mathrm{inf}}\ge0$, then
$\widetilde B_t\ge0$ because it is a convex combination.
\end{proof}

\subsection{Proof of Theorem~\ref{thm:theory-robustness}}
\label{app:proof-robustness}

We use the elementary fact that the hinge function is $1$-Lipschitz:
\begin{equation}
\left|
[x]_+-[y]_+
\right|
\le
|x-y|,
\qquad
\forall x,y\in\mathbb{R}.
\label{eq:app-hinge-lipschitz}
\end{equation}

\begin{proof}
For any fixed policy $\pi$,
\begin{align}
\left|
\mathcal{G}_{\tau_{0:T-1}}(\pi)
-
\widehat{\mathcal{G}}_{\widetilde{\tau}_{0:T-1}}(\pi)
\right|
&\le
\mathbb{E}_{\pi,\widetilde{\tau}}
\left[
\sum_{t=0}^{T-1}
\gamma^t
\left|
(
\boldsymbol{\omega}_t
-
\widetilde{\boldsymbol{\omega}}_t
)^\top
\mathbf{r}_t
\right|
\right]
\nonumber\\
&\quad+
\mathbb{E}_{\pi,\widetilde{\tau}}
\left[
\sum_{t=0}^{T-1}
\gamma^t
\lambda_\psi(\widetilde B_t,\mathbf{z}_t)
\left|
[
\bar c_{\pi,t}-B_t
]_+
-
[
\widehat c_{\pi,t}-\widetilde B_t
]_+
\right|
\right].
\end{align}
By Cauchy-Schwarz and bounded rewards, $\left|
(
\boldsymbol{\omega}_t
-
\widetilde{\boldsymbol{\omega}}_t
)^\top
\mathbf{r}_t
\right|
\le
R_2
\left\|
\boldsymbol{\omega}_t-\widetilde{\boldsymbol{\omega}}_t
\right\|_2
=
R_2\delta_t^\omega$. Using Eq.~\eqref{eq:app-hinge-lipschitz},
\begin{align}
\left|
[
\bar c_{\pi,t}-B_t
]_+
-
[
\widehat c_{\pi,t}-\widetilde B_t
]_+
\right|
&\le
\left|
(
\bar c_{\pi,t}-B_t
)
-
(
\widehat c_{\pi,t}-\widetilde B_t
)
\right|
\nonumber\\
&\le
\left|
\bar c_{\pi,t}-\widehat c_{\pi,t}
\right|
+
\left|
\widetilde B_t-B_t
\right|
=
\epsilon_t^c(\pi)+\delta_t^B .
\end{align}
Since
$\lambda_\psi(\widetilde B_t,\mathbf{z}_t)\le\Lambda$, combining the two
bounds gives Eq.~\eqref{eq:theory-robustness-bound}.

For the near-optimality bound, fix any comparator $\pi\in\Pi$ and decompose the
true performance gap through the fused model-based objective:
\begin{align}
&
\mathcal{G}_{\tau_{0:T-1}}(\pi)
-
\mathcal{G}_{\tau_{0:T-1}}(\widetilde{\pi})
\nonumber\\
&=
\left[
\mathcal{G}_{\tau_{0:T-1}}(\pi)
-
\widehat{\mathcal{G}}_{\widetilde{\tau}_{0:T-1}}(\pi)
\right]
+
\left[
\widehat{\mathcal{G}}_{\widetilde{\tau}_{0:T-1}}(\pi)
-
\widehat{\mathcal{G}}_{\widetilde{\tau}_{0:T-1}}(\widetilde{\pi})
\right] +
\left[
\widehat{\mathcal{G}}_{\widetilde{\tau}_{0:T-1}}(\widetilde{\pi})
-
\mathcal{G}_{\tau_{0:T-1}}(\widetilde{\pi})
\right].
\end{align}

The assumed $\epsilon_{\mathrm{opt}}$-optimality of $\widetilde{\pi}$ bounds the
middle term by $\epsilon_{\mathrm{opt}}$, and
Eq.~\eqref{eq:theory-robustness-bound} controls the two perturbation terms.
Taking the supremum over $\pi\in\Pi$ and using
$\epsilon_t^c(\pi)\le\bar{\epsilon}_t^c$ yields
Eq.~\eqref{eq:theory-regret-bound}.
\end{proof}

\subsection{Proof of Theorem~\ref{thm:theory-history}}
\label{app:proof-history}
We first quantify how task drift makes an $L$-step history stale relative to the
current task.

\begin{lemma}
\label{lem:app-average-staleness}
Under the drift model,
\begin{equation}
\frac{1}{L}
\sum_{\ell=1}^{L}
\left\|
\boldsymbol{\omega}_{t}
-
\boldsymbol{\omega}_{t-\ell}
\right\|_2
\le
\frac{L+1}{2}\sqrt{2}\rho_\omega,
\label{eq:app-average-pref-staleness}
\end{equation}
and
\begin{equation}
\frac{1}{L}
\sum_{\ell=1}^{L}
\left|
B_t-B_{t-\ell}
\right|
\le
\frac{L+1}{2}D_B\rho_B .
\label{eq:app-average-budget-staleness}
\end{equation}
\end{lemma}

\begin{proof}
The Euclidean diameter of the probability simplex is $\sqrt{2}$. Since Euclidean
projection onto a closed convex set is non-expansive,
\begin{align}
\left\|
\boldsymbol{\omega}_{k+1}
-
\boldsymbol{\omega}_{k}
\right\|_2
=
\left\|
\Pi_\Omega
\left(
(1-\rho_\omega)\boldsymbol{\omega}_{k}
+
\rho_\omega\bar{\boldsymbol{\omega}}_{k}
\right)
-
\Pi_\Omega(\boldsymbol{\omega}_{k})
\right\|_2\le
\rho_\omega
\left\|
\bar{\boldsymbol{\omega}}_{k}
-
\boldsymbol{\omega}_{k}
\right\|_2
\le
\sqrt{2}\rho_\omega .
\end{align}
Thus
$\|\boldsymbol{\omega}_{t}-\boldsymbol{\omega}_{t-\ell}\|_2
\le \ell\sqrt{2}\rho_\omega$. Averaging over
$\ell=1,\ldots,L$ gives Eq.~\eqref{eq:app-average-pref-staleness}. Similarly,
\begin{equation}
|B_{k+1}-B_k|
=
\rho_B|\bar B_k-B_k|
\le
D_B\rho_B ,
\end{equation}
so
$|B_t-B_{t-\ell}|\le \ell D_B\rho_B$, and averaging gives
Eq.~\eqref{eq:app-average-budget-staleness}.
\end{proof}

\begin{proof}[Proof of Theorem~\ref{thm:theory-history}]
If the task were fixed throughout the history window, the decoder errors would
scale as $A_\omega L^{-1/2}$ for preferences and $A_B L^{-1/2}$ for budgets by
assumption. Under task drift, however, the history window contains samples
generated under past tasks $\tau_{t-L},\ldots,\tau_{t-1}$ rather than the current
task $\tau_t$. Thus, the decoder error decomposes into a statistical term and a
staleness term. By the triangle inequality and
Lemma~\ref{lem:app-average-staleness},
\begin{align}
\mathbb{E}
\left[
\left\|
\boldsymbol{\omega}^{\mathrm{inf}}_t-\boldsymbol{\omega}_t
\right\|_2
\right] \le
A_\omega L^{-1/2}
+
\frac{1}{L}
\sum_{\ell=1}^{L}
\left\|
\boldsymbol{\omega}_{t-\ell}
-
\boldsymbol{\omega}_t
\right\|_2 \le
A_\omega L^{-1/2}
+
\frac{L+1}{2}\sqrt{2}\rho_\omega .
\end{align}
The budget bound follows identically
\begin{equation}
\mathbb{E}
\left[
\left|
B^{\mathrm{inf}}_t-B_t
\right|
\right]
\le
A_B L^{-1/2}
+
\frac{L+1}{2}D_B\rho_B .
\end{equation}

Ignoring lower-order constants, the preference error has the form
\begin{equation}
f(L)=A_\omega L^{-1/2}+C_\omega\rho_\omega L,
\qquad
C_\omega=\frac{\sqrt{2}}{2}.
\end{equation}
Treating $L$ as continuous,
\begin{equation}
f'(L)
=
-\frac{1}{2}A_\omega L^{-3/2}
+
C_\omega\rho_\omega .
\end{equation}
Setting $f'(L)=0$ yields
\begin{equation}
L_\omega^\star
=
\left(
\frac{A_\omega}{2C_\omega\rho_\omega}
\right)^{2/3}
\asymp
\left(
\frac{A_\omega}{\rho_\omega}
\right)^{2/3}.
\end{equation}
Substituting this value into $f(L)$ gives
\begin{equation}
\min_L f(L)
=
\mathcal{O}
\left(
A_\omega^{2/3}\rho_\omega^{1/3}
\right).
\end{equation}
The budget result follows by replacing $A_\omega$ with $A_B$ and
$C_\omega\rho_\omega$ with $(D_B/2)\rho_B$.
\end{proof}

\subsection{Oracle Fusion and Inverse-Uncertainty Weighting}
\label{app:oracle-fusion}

The uncertainty gate in \algname\ can be interpreted as an amortized
approximation to an oracle fusion rule. For a generic task variable $\theta_t$,
let $\widehat{\theta}_t$ denote the exposed estimate and
$\theta_t^{\mathrm{inf}}$ the belief-derived estimate. We consider the gated
fusion
\begin{equation}
\widetilde{\theta}_t(\alpha)
=
(1-\alpha)\widehat{\theta}_t
+
\alpha\theta_t^{\mathrm{inf}},
\qquad
\alpha\in[0,1].
\end{equation}

\begin{proposition}
\label{prop:app-oracle-fusion}
Let
$e^{\mathrm{exp}}=\widehat{\theta}_t-\theta_t$ and
$e^{\mathrm{inf}}=\theta_t^{\mathrm{inf}}-\theta_t$. The mean-squared fusion
risk $\mathcal{R}(\alpha)
=
\mathbb{E}
\left[
\left\|
(1-\alpha)e^{\mathrm{exp}}
+
\alpha e^{\mathrm{inf}}
\right\|_2^2
\right]$ is minimized over $\alpha\in[0,1]$ by
\begin{equation}
\alpha^\star
=
\Pi_{[0,1]}
\left(
\frac{
\mathbb{E}\|e^{\mathrm{exp}}\|_2^2
-
\mathbb{E}\langle e^{\mathrm{exp}},e^{\mathrm{inf}}\rangle
}{
\mathbb{E}\|e^{\mathrm{exp}}-e^{\mathrm{inf}}\|_2^2
}
\right).
\label{eq:app-oracle-alpha}
\end{equation}
If the two errors are unbiased, independent, and have variances
$\sigma_{\mathrm{exp}}^2$ and $\sigma_{\mathrm{inf}}^2$, then
\begin{equation}
\alpha^\star
=
\frac{
\sigma_{\mathrm{exp}}^2
}{
\sigma_{\mathrm{exp}}^2+\sigma_{\mathrm{inf}}^2
}.
\label{eq:app-inverse-variance}
\end{equation}
\end{proposition}

\begin{proof}
Let $d=e^{\mathrm{inf}}-e^{\mathrm{exp}}$. The fusion error can be written as
$e^{\mathrm{exp}}+\alpha d$, hence
\begin{align}
\mathcal{R}(\alpha)
&=
\mathbb{E}
\left[
\left\|
e^{\mathrm{exp}}+\alpha d
\right\|_2^2
\right] =
\mathbb{E}\|e^{\mathrm{exp}}\|_2^2
+
2\alpha
\mathbb{E}\langle e^{\mathrm{exp}},d\rangle
+
\alpha^2
\mathbb{E}\|d\|_2^2 .
\end{align}
This is a convex quadratic. The unconstrained minimizer is
\begin{equation}
\alpha_{\mathrm{unc}}^\star
=
-
\frac{
\mathbb{E}\langle e^{\mathrm{exp}},d\rangle
}{
\mathbb{E}\|d\|_2^2
}
=
\frac{
\mathbb{E}\|e^{\mathrm{exp}}\|_2^2
-
\mathbb{E}\langle e^{\mathrm{exp}},e^{\mathrm{inf}}\rangle
}{
\mathbb{E}\|e^{\mathrm{exp}}-e^{\mathrm{inf}}\|_2^2
}.
\end{equation}
Projecting onto $[0,1]$ gives Eq.~\eqref{eq:app-oracle-alpha}. Under
unbiasedness and independence,
$\mathbb{E}\langle e^{\mathrm{exp}},e^{\mathrm{inf}}\rangle=0$, which gives
Eq.~\eqref{eq:app-inverse-variance}.
\end{proof}

\subsection{Posterior Uncertainty and Decoded-Task Variance}
\label{app:decoder-uncertainty}
Let
\begin{equation}
d_\omega(\mathbf{z})
=
\frac{
\varphi(g_\omega(\mathbf{z}))+\varepsilon\mathbf{1}_{d_\omega}
}{
\mathbf{1}_{d_\omega}^{\top}
\left(
\varphi(g_\omega(\mathbf{z}))+\varepsilon\mathbf{1}_{d_\omega}
\right)
},
\qquad
d_B(\mathbf{z})
=
\varphi(g_B(\mathbf{z})),
\label{eq:app-decoders}
\end{equation}
where $\varphi(x)=\log(1+\exp(x))$, 
$\mathbf{1}_{d_\omega}\in\mathbb{R}^{d_\omega}$ denotes the all-ones vector, 
$d_\omega$ is the preference dimension, and $\varepsilon>0$ ensures strictly positive normalized preference weights.

\begin{lemma}
\label{lem:app-decoder-uncertainty}
Assume $g_\omega$ and $g_B$ are Lipschitz with constants
$L_{g_\omega}$ and $L_{g_B}$. Then $d_B$ is
$L_{g_B}$-Lipschitz. Moreover, $d_\omega$ is Lipschitz with constant
\begin{equation}
L_\omega
\le
\frac{\sqrt{K(K+3)}}{K\varepsilon}
L_{g_\omega}.
\label{eq:app-pref-lipschitz}
\end{equation}
Therefore, if
$\mathbf{z}_t\sim
\mathcal{N}(\boldsymbol{\mu}_t,\mathrm{Diag}(\boldsymbol{\sigma}_t^2))$, then
\begin{equation}
\mathbb{E}
\left[
\left\|
d_\omega(\mathbf{z}_t)-d_\omega(\boldsymbol{\mu}_t)
\right\|_2^2
\mid H_t
\right]
\le
L_\omega^2
\|\boldsymbol{\sigma}_t\|_2^2,
\label{eq:app-pref-var-bound}
\end{equation}
and
\begin{equation}
\mathbb{E}
\left[
\left|
d_B(\mathbf{z}_t)-d_B(\boldsymbol{\mu}_t)
\right|^2
\mid H_t
\right]
\le
L_{g_B}^2
\|\boldsymbol{\sigma}_t\|_2^2.
\label{eq:app-budget-var-bound}
\end{equation}
\end{lemma}
This lemma shows that posterior variance controls the sensitivity of decoded task estimates to latent
sampling noise, supporting its use as a practical reliability proxy.

\subsection{Dual Interpretation for Instantaneous Safety}
\label{app:dual-interpretation}

The context-dependent dual variable in \algname\ is
\begin{equation}
\lambda_\psi(\widetilde B_t,\mathbf{z}_t)
=
\Lambda \sigma(f_\lambda(\widetilde B_t,\mathbf{z}_t)),
\end{equation}
where $\sigma(\cdot)$ is the sigmoid function. This parameterization ensures
$0\le\lambda_\psi(\widetilde B_t,\mathbf{z}_t)\le\Lambda$ and preserves
monotonicity with respect to the pre-activation $f_\lambda$. For one sample, the dual loss is
\begin{equation}
\ell_\lambda(\psi)
=
-
\lambda_\psi(\widetilde B_t,\mathbf{z}_t)
\left(
\widehat c_t(\mathbf{a}_t)-\widetilde B_t
\right).
\label{eq:app-single-dual-loss}
\end{equation}
Its gradient is
\begin{equation}
\nabla_\psi \ell_\lambda(\psi)
=
-
\Lambda
\left(
\widehat c_t(\mathbf{a}_t)-\widetilde B_t
\right)
\sigma(f_\lambda(\widetilde B_t,\mathbf{z}_t))
\left(
1-\sigma(f_\lambda(\widetilde B_t,\mathbf{z}_t))
\right)
\nabla_\psi f_\lambda(\widetilde B_t,\mathbf{z}_t).
\label{eq:app-dual-gradient}
\end{equation}
Thus, gradient descent on the dual loss increases the dual pre-activation when
the predicted instantaneous violation
$\widehat c_t(\mathbf{a}_t)-\widetilde B_t$ is positive, strengthening the actor
penalty. When the predicted violation is negative, the update reduces the dual
penalty and avoids excessive conservatism. The gap between predicted and true instantaneous safety is bounded by the
cost-model error and the budget-fusion error, as characterized in
Theorem~\ref{thm:theory-robustness}.

\section{Environment Setup}
\label{app:env_setup}

We evaluate \algname{} on multi-objective MuJoCo locomotion tasks and a
multi-zone building-control task. Each environment returns a vector reward
$\mathbf{r}_t\in\mathbb{R}^{K}$, and a preference
$\boldsymbol{\omega}\in\Omega$ scalarizes the return as
$\boldsymbol{\omega}^{\top}\sum_t\mathbf{r}_t$. Feasibility is controlled by an
external budget channel rather than by adding cost as an extra objective. Unless
otherwise stated, the per-step cost is
$c_t=0.01K_a^{-1}\sum_{i=1}^{K_a}a_{t,i}^{2}$, where $K_a$ is the action
dimension, and a trajectory is feasible when its average cost satisfies
$T^{-1}\sum_{t=0}^{T-1}c_t\le B$ with $B=0.01$. For static feasible-front evaluation, $B$ is fixed and preferences are drawn
from the evaluation grid. For task-drift evaluation, the hidden preference and
budget are refreshed every $50$ steps. We use $\rho_\omega=0.25$ for preference
drift, $\rho_B=0.30$ for budget drift, $B_t\in[0.005,0.020]$, preference
observation noise $0.15$, and budget observation delay $20$ steps.

\begin{table}[t]
\centering
\caption{Environment and drift settings}
\label{tab:environment_summary}
\scriptsize
\setlength{\tabcolsep}{3pt}
\renewcommand{\arraystretch}{1.12}
\begin{adjustbox}{width=0.8\textwidth}
\begin{tabular}{lcccccccc}
\toprule
\textbf{Environment}
& \textbf{$K$}
& \textbf{State dim.}
& \textbf{Action dim.}
& \makecell{\textbf{Eval.}\\\textbf{prefs.}}
& \textbf{$B$}
& \textbf{$\rho_\omega$}
& \textbf{$\rho_B$}
& \makecell{\textbf{Drift}\\\textbf{$B_t$ range}} \\
\midrule
MO-Ant-2d              & 2 & 27  & 8  & 201 & 0.01 & 0.25 & 0.30 & $[0.005,0.020]$ \\
MO-HalfCheetah-2d      & 2 & 17  & 6  & 201 & 0.01 & 0.25 & 0.30 & $[0.005,0.020]$ \\
MO-Hopper-2d           & 2 & 11  & 3  & 201 & 0.01 & 0.25 & 0.30 & $[0.005,0.020]$ \\
MO-Humanoid            & 2 & 376 & 17 & 201 & 0.01 & 0.25 & 0.30 & $[0.005,0.020]$ \\
MO-Ant-3d              & 3 & 27  & 8  & 231 & 0.01 & 0.25 & 0.30 & $[0.005,0.020]$ \\
MO-Ant-4d              & 4 & 27  & 8  & 286 & 0.01 & 0.25 & 0.30 & $[0.005,0.020]$ \\
MO-Ant-5d              & 5 & 27  & 8  & 126 & 0.01 & 0.25 & 0.30 & $[0.005,0.020]$ \\
MO-HalfCheetah-5d      & 5 & 17  & 6  & 126 & 0.01 & 0.25 & 0.30 & $[0.005,0.020]$ \\
Building-9d            & 9 & 29  & 23 & 32  & 0.01 & 0.25 & 0.30 & $[0.005,0.020]$ \\
\bottomrule
\end{tabular}
\end{adjustbox}
\end{table}

\begin{itemize}
    \item \textbf{MO-HalfCheetah-2d.}
    The environment has $\mathcal{S}\subset\mathbb{R}^{17}$,
    $\mathcal{A}\subset\mathbb{R}^{6}$, and $K=2$. Let $v_x$ denote the forward
    velocity and $C_{\mathrm{alive}}=1$. The objectives are
    $R_1=\min(4,v_x)+C_{\mathrm{alive}}$ and
    $R_2=4-\sum_i a_i^2+C_{\mathrm{alive}}$, corresponding to forward speed and
    energy-efficient control.

    \item \textbf{MO-HalfCheetah-5d.}
    This task uses the same state and action dimensions as MO-HalfCheetah-2d,
    with $K=5$. Let $h$ be the body height and $h_0$ the initial body height.
    The objectives are $R_1=0.5v_x+C_{\mathrm{alive}}$,
    $R_2=20(h-h_0)+C_{\mathrm{alive}}$,
    $R_3=4-\frac{3}{2}(a_0^2+a_3^2)+C_{\mathrm{alive}}$,
    $R_4=4-\frac{3}{2}(a_1^2+a_2^2)+C_{\mathrm{alive}}$, and
    $R_5=4-\frac{3}{2}(a_3^2+a_4^2)+C_{\mathrm{alive}}$. These objectives
    capture forward speed, jump height, and actuator-group efficiency.

    \item \textbf{MO-Ant-2d.}
    The environment has $\mathcal{S}\subset\mathbb{R}^{27}$,
    $\mathcal{A}\subset\mathbb{R}^{8}$, and $K=2$. Let $v_x$ and $v_y$ denote
    torso velocities along the two horizontal axes. With
    $R_{\mathrm{other}}=1-0.5\sum_i a_i^2$, the objectives are
    $R_1=v_x+R_{\mathrm{other}}$ and $R_2=v_y+R_{\mathrm{other}}$, encouraging
    locomotion in two directions while penalizing large actions.

    \item \textbf{MO-Ant-3d.}
    This task uses $\mathcal{S}\subset\mathbb{R}^{27}$,
    $\mathcal{A}\subset\mathbb{R}^{8}$, and $K=3$. The objectives are
    $R_1=v_x$, $R_2=v_y$, and $R_3=4-0.5\sum_i a_i^2$, corresponding to
    movement along two horizontal axes and energy efficiency.

    \item \textbf{MO-Ant-4d.}
    This task uses the Ant state and action spaces with $K=4$. Each objective
    combines forward velocity with one actuator-pair efficiency term:
    $R_1=4-2(a_0^2+a_1^2)+v_x$,
    $R_2=4-2(a_2^2+a_3^2)+v_x$,
    $R_3=4-2(a_4^2+a_5^2)+v_x$, and
    $R_4=4-2(a_6^2+a_7^2)+v_x$.

    \item \textbf{MO-Ant-5d.}
    This task separates forward speed from four actuator-group efficiency
    objectives: $R_1=v_x$, $R_2=4-2(a_0^2+a_1^2)$,
    $R_3=4-2(a_2^2+a_3^2)$, $R_4=4-2(a_4^2+a_5^2)$, and
    $R_5=4-2(a_6^2+a_7^2)$. Training preferences are sampled hierarchically by
    first allocating mass between speed and energy and then distributing the
    energy mass across actuator groups.

    \item \textbf{MO-Hopper-2d.}
    The environment has $\mathcal{S}\subset\mathbb{R}^{11}$,
    $\mathcal{A}\subset\mathbb{R}^{3}$, and $K=2$. Let $v_x$ denote forward
    velocity, $h$ the body height, and $h_0$ the initial height. With
    $R_{\mathrm{other}}=1-2\times10^{-4}\sum_i a_i^2$, the objectives are
    $R_1=1.5v_x+R_{\mathrm{other}}$ and
    $R_2=12(h-h_0)+R_{\mathrm{other}}$, rewarding forward running and jumping
    height. Since Hopper actions have a larger environment-scale range than
    $[-1,1]$, the action-cost constraint is more demanding.

    \item \textbf{MO-Humanoid.}
    The environment has $\mathcal{S}\subset\mathbb{R}^{376}$,
    $\mathcal{A}\subset\mathbb{R}^{17}$, and $K=2$. Let $v_x$ denote
    mass-center velocity and $C_{\mathrm{alive}}=3$. The objectives are
    $R_1=1.25v_x+C_{\mathrm{alive}}$ and
    $R_2=3-4\sum_i \mathrm{ctrl}_i^2+C_{\mathrm{alive}}$, capturing forward
    locomotion and energy-efficient control. Episodes terminate when the torso
    height leaves $[1.0,2.0]$.

    \item \textbf{Building-9d.}
    The building-control task uses the OfficeLarge building under Warm Marine
    weather in ElPaso. The environment has $\mathcal{S}\subset\mathbb{R}^{29}$,
    $\mathcal{A}\subset\mathbb{R}^{23}$, and $K=9$. The action dimensions
    correspond to HVAC controls for 23 zones, and observations include
    normalized zone temperatures and exogenous variables such as outdoor
    temperature, irradiance, ground temperature, occupancy power, carbon
    intensity, and electricity price. The zones are partitioned into bottom,
    middle, and top groups with sizes $9$, $7$, and $7$. Let
    $e=X_{\mathrm{new}}\odot A_{\mathrm{C}}-20A_{\mathrm{C}}$ be the
    temperature error relative to $20^\circ\mathrm{C}$. The comfort objectives
    are $R_1=(20\cdot9-\|e_{\mathrm{bottom}}\|_1)/20$,
    $R_2=(20\cdot7-\|e_{\mathrm{middle}}\|_1)/20$, and
    $R_3=(20\cdot7-\|e_{\mathrm{top}}\|_1)/20$. The energy objectives are
    $R_4=(20\cdot9-\|a_{\mathrm{bottom}}\|_1)/20$,
    $R_5=(20\cdot7-\|a_{\mathrm{middle}}\|_1)/20$, and
    $R_6=(20\cdot7-\|a_{\mathrm{top}}\|_1)/20$. The ramping objectives are
    $R_7=9-\left|\|a_{\mathrm{bottom}}^{t-1}\|_1-\|a_{\mathrm{bottom}}^{t-2}\|_1\right|$,
    $R_8=7-\left|\|a_{\mathrm{middle}}^{t-1}\|_1-\|a_{\mathrm{middle}}^{t-2}\|_1\right|$, and
    $R_9=7-\left|\|a_{\mathrm{top}}^{t-1}\|_1-\|a_{\mathrm{top}}^{t-2}\|_1\right|$.
    For the first two transitions, the ramping rewards are set to zero.
\end{itemize}

\section{Implementation Details}
\label{app:implementation_details}

\algname{} uses a shared latent state encoder, a history-conditioned belief
encoder, twin distributional critics, a tanh-squashed Gaussian actor, a
distributional cost model, and a context-dependent dual network. The state
encoder produces a 64-dimensional latent embedding, and the GRU belief encoder
infers a 64-dimensional latent task context from recent transition features
$[\mathbf{s}_t,\mathbf{a}_t,\mathbf{r}_t,\widehat{\boldsymbol{\omega}}_t,c_t,
\widehat B_t]$. All experiments are run on a workstation with an Intel Core i9-14900K CPU, 64GB
RAM, and an NVIDIA GeForce RTX 5070 Ti GPU. The main implementation parameters
are summarized in Table~\ref{tab:becon_implementation_details}.

\begin{table}[!ht]
\centering
\caption{\algname{} implementation details}
\label{tab:becon_implementation_details}
\scriptsize
\setlength{\tabcolsep}{6pt}
\renewcommand{\arraystretch}{1.12}
\begin{adjustbox}{width=0.75\textwidth}
\begin{tabular}{ll}
\toprule
\textbf{Parameter} & \textbf{Setting} \\
\midrule
State encoder & MLP $(256,256)$, ELU, LayerNorm output \\
Latent dimension & $64$ \\
Belief dimension & $64$ \\
Actor & Tanh-squashed Gaussian policy, MLP $(256,256)$ \\
Critics & Twin distributional critics with $51$ atoms per objective \\
Value support & Initial support $[-500,500]$, adaptive support enabled \\
Cost model & Distributional cost head with $51$ atoms on $[0,1]$ \\
Dual variable & $\lambda_\psi(\widetilde B_t,\mathbf{z}_t)$ \\
Optimizer & Adam \\
Replay buffer size & $10^6$ transitions \\
Batch size & $256$ \\
Discount factor & $\gamma=0.99$ \\
Target update rate & $0.005$ \\
Learning rates & $3\times10^{-4}$ for actor/main networks; $10^{-4}$ for dual \\
Entropy coefficient & $\alpha_H=0.2$ \\
\bottomrule
\end{tabular}
\end{adjustbox}
\end{table}

\section{Additional Experiments}
\label{app:additional_experiments}

\subsection{Constraint-Tightness Sensitivity}
\label{app:budget_sensitivity}

We further evaluate trained policies under stringent admissible cost levels to
test whether feasible-front quality is preserved when the stage-wise constraint
becomes tighter. Table~\ref{tab:budget_sensitivity} shows that BECON maintains
low violation rates and nonzero feasible hypervolume in cases where COLA becomes
partially or fully infeasible.

\begin{table}[t]
\centering
\caption{Constraint-tightness sensitivity under different admissible cost levels}
\label{tab:budget_sensitivity}
\begin{tabular}{llccc}
\toprule
Dataset & Method & $B$ & VR $\downarrow$ & FHV $\uparrow$ \\
\midrule
Building-9d & BECON & 0.005 & 0.000 & $6.82{\times}10^{31}$ \\
Building-9d & COLA  & 0.005 & 1.000 & 0 \\
MO-Hopper-2d & BECON & 0.010 & 0.005 & $1.29{\times}10^{7}$ \\
MO-Hopper-2d & COLA  & 0.010 & 1.000 & 0 \\
MO-HalfCheetah-5d & BECON & 0.005 & 0.000 & $1.15{\times}10^{17}$ \\
MO-HalfCheetah-5d & COLA  & 0.005 & 0.548 & $3.04{\times}10^{16}$ \\
\bottomrule
\end{tabular}
\end{table}

\subsection{Component Ablation}
\label{app:component_ablation}

\begin{table}[!ht]
\centering
\caption{Component ablation under feasibility constraints}
\label{tab:component_ablation_feasible}
\scriptsize
\setlength{\tabcolsep}{4.8pt}
\renewcommand{\arraystretch}{1.12}
\begin{adjustbox}{width=\textwidth}
\begin{tabular}{llcccc}
\toprule
\textbf{Dataset} & \textbf{Variant}
& \textbf{FHV} $\uparrow$
& \textbf{FUT} $\uparrow$
& \makecell{\textbf{MC}\\$(\times 10^{-3})$ $\downarrow$}
& $|\mathcal{P}|$ $\uparrow$ \\
\midrule

\multirow{4}{*}{MO-Ant-2d}
& Full \algname{}
& $\mathbf{5.790\times 10^{6}}$
& $\mathbf{2224.97}$
& $1.548$
& $\mathbf{45}$ \\
& Exposed descriptor only
& $5.695\times 10^{6}$
& $2224.55$
& $1.634$
& $39$ \\
& w/o constraint penalty
& $5.633\times 10^{6}$
& $2203.82$
& $1.603$
& $34$ \\
& w/o history belief
& $5.115\times 10^{6}$
& $2100.43$
& $\mathbf{1.473}$
& $38$ \\
\midrule

\multirow{4}{*}{MO-HalfCheetah-2d}
& Full \algname{}
& $\mathbf{5.789\times 10^{6}}$
& $\mathbf{2346.69}$
& $\mathbf{0.806}$
& $80$ \\
& Exposed descriptor only
& $5.675\times 10^{6}$
& $2342.06$
& $0.924$
& $68$ \\
& w/o constraint penalty
& $5.737\times 10^{6}$
& $2340.16$
& $0.882$
& $\mathbf{92}$ \\
& w/o history belief
& $5.648\times 10^{6}$
& $2341.94$
& $0.924$
& $82$ \\
\midrule

\multirow{4}{*}{MO-Ant-3d}
& Full \algname{}
& $\mathbf{7.377\times 10^{9}}$
& $\mathbf{1823.95}$
& $1.364$
& $\mathbf{134}$ \\
& Exposed descriptor only
& $6.163\times 10^{9}$
& $1712.09$
& $1.970$
& $116$ \\
& w/o constraint penalty
& $7.032\times 10^{9}$
& $1797.64$
& $\mathbf{1.285}$
& $106$ \\
& w/o history belief
& $6.629\times 10^{9}$
& $1758.62$
& $1.888$
& $110$ \\
\midrule

\multirow{4}{*}{MO-Ant-4d}
& Full \algname{}
& $\mathbf{3.739\times 10^{14}}$
& $\mathbf{4380.32}$
& $1.529$
& $\mathbf{17}$ \\
& Exposed descriptor only
& $3.561\times 10^{14}$
& $4335.54$
& $1.659$
& $15$ \\
& w/o constraint penalty
& $2.942\times 10^{14}$
& $4123.29$
& $1.866$
& $11$ \\
& w/o history belief
& $3.374\times 10^{14}$
& $4276.89$
& $\mathbf{1.483}$
& $4$ \\
\midrule

\multirow{4}{*}{MO-Ant-5d}
& Full \algname{}
& $\mathbf{3.013\times 10^{16}}$
& $\mathbf{1991.09}$
& $0.255$
& $97$ \\
& Exposed descriptor only
& $2.594\times 10^{16}$
& $1906.18$
& $0.651$
& $90$ \\
& w/o constraint penalty
& $2.563\times 10^{16}$
& $1897.47$
& $\mathbf{0.239}$
& $94$ \\
& w/o history belief
& $2.400\times 10^{16}$
& $1889.73$
& $0.531$
& $\mathbf{104}$ \\
\bottomrule
\end{tabular}
\end{adjustbox}
\end{table}

Table~\ref{tab:component_ablation_feasible} isolates the contribution of the
main components of \algname{}. The full model achieves the best FHV and FUT
across all tasks, indicating that history-conditioned belief inference and
constraint-aware control jointly improve feasible-front quality. Using only the
exposed descriptor consistently reduces FHV, especially on higher-dimensional
Ant tasks, showing that recent histories provide useful evidence for recovering
the latent task context. Removing the constraint penalty can reduce MC in some
cases, but this does not translate into better feasible-front quality, as FHV,
FUT, and $|\mathcal{P}|$ often decrease. These results suggest that \algname{}
improves constrained MORL performance by balancing preference-aligned return
with feasibility, rather than by merely minimizing cost.

\subsection{History Encoder Architecture}
\label{app:history_encoder_ablation}

\begin{table}[!ht]
\centering
\caption{History-encoder ablation under task drift}
\label{tab:architecture_drift_ablation}
\scriptsize
\setlength{\tabcolsep}{4.8pt}
\renewcommand{\arraystretch}{1.12}
\begin{adjustbox}{width=0.9\textwidth}
\begin{tabular}{llccccc}
\toprule
\textbf{Dataset} & \textbf{Encoder}
& \textbf{FHV} $\uparrow$
& \textbf{FUT} $\uparrow$
& \makecell{\textbf{MC}\\$(\times 10^{-3})$ $\downarrow$}
& \textbf{SR} $\uparrow$
& \textbf{PCU} $\uparrow$ \\
\midrule

\multirow{4}{*}{MO-HalfCheetah-2d}
& GRU
& $\mathbf{5.797\times 10^{6}}$
& $\mathbf{2351.60}$
& $\mathbf{0.793}$
& $\mathbf{4.437}$
& $\mathbf{4.366}$ \\
& MLP
& $5.673\times 10^{6}$
& $2348.11$
& $0.797$
& $4.421$
& $4.356$ \\
& TCN
& $5.778\times 10^{6}$
& $2344.25$
& $0.801$
& $4.428$
& $4.350$ \\
& Transformer
& $5.775\times 10^{6}$
& $2345.66$
& $0.820$
& $4.413$
& $4.360$ \\
\midrule

\multirow{4}{*}{MO-Ant-4d}
& GRU
& $\mathbf{4.418\times 10^{14}}$
& $\mathbf{4564.29}$
& $1.624$
& $\mathbf{9.064}$
& $\mathbf{9.198}$ \\
& MLP
& $4.189\times 10^{14}$
& $4489.59$
& $\mathbf{1.558}$
& $8.951$
& $9.073$ \\
& TCN
& $4.103\times 10^{14}$
& $4464.43$
& $1.604$
& $8.660$
& $8.868$ \\
& Transformer
& $4.351\times 10^{14}$
& $4545.02$
& $1.647$
& $8.999$
& $9.059$ \\
\midrule

\multirow{4}{*}{MO-Ant-5d}
& GRU
& $3.099\times 10^{16}$
& $1976.85$
& $\mathbf{0.292}$
& $3.862$
& $3.866$ \\
& MLP
& $2.804\times 10^{16}$
& $1942.19$
& $0.340$
& $3.725$
& $3.719$ \\
& TCN
& $\mathbf{3.259\times 10^{16}}$
& $2004.75$
& $0.359$
& $3.969$
& $4.040$ \\
& Transformer
& $3.247\times 10^{16}$
& $\mathbf{2013.76}$
& $0.426$
& $\mathbf{4.053}$
& $\mathbf{4.064}$ \\

\bottomrule
\end{tabular}
\end{adjustbox}
\end{table}

Table~\ref{tab:architecture_drift_ablation} evaluates different temporal
encoders for the belief module under task drift. GRU achieves the best overall
performance on MO-HalfCheetah-2d and MO-Ant-4d, with the strongest FHV, FUT, SR,
and PCU in both tasks. On MO-Ant-5d, TCN and Transformer become more competitive,
suggesting that higher-dimensional preference structure can benefit from richer
temporal aggregation. Overall, the results indicate that \algname{} is not tied
to a single encoder architecture, while the GRU provides a strong and stable
default across tasks.

\subsection{History Length Sensitivity}
\label{app:history_length_ablation}

\begin{table}[!http]
\centering
\caption{History-window ablation by dataset}
\label{tab:history_length_by_dataset}
\scriptsize
\setlength{\tabcolsep}{4.2pt}
\renewcommand{\arraystretch}{1.12}
\begin{adjustbox}{width=0.8\textwidth}
\begin{tabular}{llccccc}
\toprule
\textbf{Dataset} & \textbf{$L$}
& \textbf{FHV} $\uparrow$
& \textbf{FUT} $\uparrow$
& \makecell{\textbf{MC}\\$(\times 10^{-3})$ $\downarrow$}
& \textbf{SR} $\uparrow$
& \textbf{PCU} $\uparrow$ \\
\midrule

\multirow{5}{*}{MO-HalfCheetah-2d}
& 0
& $5.653\times 10^{6}$
& $2339.17$
& $0.885$
& $4.404$
& $4.339$ \\
& 2
& $5.725\times 10^{6}$
& $2339.04$
& $\mathbf{0.845}$
& $4.396$
& $4.341$ \\
& 4
& $5.676\times 10^{6}$
& $\mathbf{2340.38}$
& $0.945$
& $\mathbf{4.544}$
& $\mathbf{4.563}$ \\
& 8
& $\mathbf{5.776\times 10^{6}}$
& $2332.76$
& $0.938$
& $4.413$
& $4.360$ \\
& 16
& $5.763\times 10^{6}$
& $2336.99$
& $0.931$
& $4.397$
& $4.324$ \\
\midrule

\multirow{5}{*}{MO-Ant-4d}
& 0
& $2.346\times 10^{14}$
& $3876.40$
& $1.816$
& $7.334$
& $7.515$ \\
& 2
& $3.385\times 10^{14}$
& $4270.87$
& $\mathbf{1.694}$
& $8.340$
& $8.492$ \\
& 4
& $\mathbf{3.687\times 10^{14}}$
& $\mathbf{4355.44}$
& $1.696$
& $\mathbf{8.600}$
& $\mathbf{8.651}$ \\
& 8
& $3.304\times 10^{14}$
& $4248.51$
& $1.749$
& $7.982$
& $8.285$ \\
& 16
& $3.006\times 10^{14}$
& $4142.39$
& $1.754$
& $7.943$
& $8.234$ \\
\midrule

\multirow{5}{*}{MO-HalfCheetah-5d}
& 0
& $3.325\times 10^{16}$
& $2369.11$
& $1.515$
& $5.356$
& $5.976$ \\
& 2
& $\mathbf{9.502\times 10^{16}}$
& $\mathbf{2682.32}$
& $1.473$
& $6.502$
& $6.618$ \\
& 4
& $7.574\times 10^{16}$
& $2635.58$
& $1.486$
& $\mathbf{6.981}$
& $\mathbf{11.477}$ \\
& 8
& $5.762\times 10^{16}$
& $2250.90$
& $\mathbf{1.149}$
& $4.159$
& $4.761$ \\
& 16
& $8.435\times 10^{16}$
& $2584.66$
& $1.459$
& $5.915$
& $4.491$ \\

\bottomrule
\end{tabular}
\end{adjustbox}
\end{table}

Table~\ref{tab:history_length_by_dataset} evaluates the effect of the history
window length $L$. The optimal value of $L$ is task-dependent, reflecting the
trade-off between useful temporal evidence and stale task information under
drift. On MO-HalfCheetah-2d, $L=8$ gives the best FHV, while $L=4$ gives the
best FUT, SR, and PCU, suggesting that short-to-moderate histories are sufficient
for this lower-dimensional task. On MO-Ant-4d, $L=4$ achieves the strongest FHV,
FUT, SR, and PCU, whereas longer histories degrade performance, indicating that
excessively old transitions can hurt adaptation. On MO-HalfCheetah-5d, nonzero
history is essential: $L=2$ yields the best FHV and FUT, while $L=4$ achieves
the best SR and PCU. Overall, these results support the role of history-based
belief inference while demonstrating that the history window should remain short
enough to avoid staleness under task drift.

\subsection{Uncertainty-Gated Fusion Sensitivity}
\label{app:intent_mix_ablation}

\begin{table}[!ht]
\centering
\caption{Intent-mixing ablation by dataset}
\label{tab:intent_mix_by_dataset}
\scriptsize
\setlength{\tabcolsep}{4.2pt}
\renewcommand{\arraystretch}{1.12}
\begin{adjustbox}{width=0.8\textwidth}
\begin{tabular}{llccccc}
\toprule
\textbf{Dataset} & \textbf{$\alpha_{\max}$}
& \textbf{FHV} $\uparrow$
& \textbf{FUT} $\uparrow$
& \makecell{\textbf{MC}\\$(\times 10^{-3})$ $\downarrow$}
& \textbf{SR} $\uparrow$
& \textbf{PCU} $\uparrow$ \\
\midrule

\multirow{5}{*}{MO-HalfCheetah-2d}
& 0.00
& $5.655\times 10^{6}$
& $2341.04$
& $0.877$
& $4.410$
& $4.336$ \\
& 0.25
& $5.703\times 10^{6}$
& $2342.70$
& $0.869$
& $4.492$
& $4.467$ \\
& 0.50
& $5.694\times 10^{6}$
& $\mathbf{2343.06}$
& $0.953$
& $\mathbf{4.493}$
& $\mathbf{4.459}$ \\
& 0.75
& $\mathbf{5.724\times 10^{6}}$
& $2323.67$
& $0.866$
& $4.256$
& $4.164$ \\
& 1.00
& $5.714\times 10^{6}$
& $2297.31$
& $\mathbf{0.791}$
& $3.836$
& $3.678$ \\
\midrule

\multirow{5}{*}{MO-Ant-4d}
& 0.00
& $3.657\times 10^{14}$
& $4351.41$
& $1.719$
& $8.457$
& $8.520$ \\
& 0.25
& $3.647\times 10^{14}$
& $4344.34$
& $1.863$
& $8.465$
& $8.639$ \\
& 0.50
& $3.504\times 10^{14}$
& $4300.74$
& $1.734$
& $8.447$
& $8.701$ \\
& 0.75
& $3.573\times 10^{14}$
& $4331.54$
& $1.717$
& $\mathbf{8.501}$
& $\mathbf{8.796}$ \\
& 1.00
& $\mathbf{3.684\times 10^{14}}$
& $\mathbf{4362.46}$
& $\mathbf{1.634}$
& $8.417$
& $8.582$ \\
\midrule

\multirow{5}{*}{MO-HalfCheetah-5d}
& 0.00
& $3.548\times 10^{16}$
& $2378.05$
& $1.603$
& $5.311$
& $6.072$ \\
& 0.25
& $7.427\times 10^{16}$
& $2540.92$
& $1.765$
& $\mathbf{6.935}$
& $\mathbf{11.461}$ \\
& 0.50
& $\mathbf{8.552\times 10^{16}}$
& $\mathbf{2555.41}$
& $1.474$
& $5.673$
& $5.709$ \\
& 0.75
& $7.622\times 10^{16}$
& $2418.98$
& $1.179$
& $4.806$
& $3.849$ \\
& 1.00
& $0.000$
& $1617.87$
& $\mathbf{0.210}$
& $1.657$
& $1.684$ \\

\bottomrule
\end{tabular}
\end{adjustbox}
\end{table}

Table~\ref{tab:intent_mix_by_dataset} studies the effect of the maximum fusion
weight $\alpha_{\max}$, which controls the reliance on belief-derived task
estimates. The best setting varies across datasets and metrics. On
MO-HalfCheetah-2d, moderate fusion yields the strongest online adaptation, with
$\alpha_{\max}=0.5$ achieving the best FUT, SR, and PCU, while
$\alpha_{\max}=0.75$ gives the highest FHV and $\alpha_{\max}=1.0$ obtains the
lowest MC. On MO-Ant-4d, stronger belief reliance is beneficial for static
feasible-front quality and cost control, with $\alpha_{\max}=1.0$ achieving the
best FHV, FUT, and MC, whereas $\alpha_{\max}=0.75$ gives the best SR and PCU.
On MO-HalfCheetah-5d, intermediate fusion is crucial: $\alpha_{\max}=0.5$
achieves the best FHV and FUT, while $\alpha_{\max}=0.25$ gives the strongest SR
and PCU. Fully relying on inferred task signals can be harmful, as
$\alpha_{\max}=1.0$ collapses FHV on MO-HalfCheetah-5d despite producing the
lowest MC. These results support uncertainty-gated fusion: belief-derived
estimates improve adaptation and feasible-front quality, but should be balanced
with the exposed descriptor to avoid overly conservative or miscalibrated task
conditioning.

\subsection{Preference Drift Sensitivity}
\label{app:preference_drift_ablation}

\begin{table}[!ht]
\centering
\caption{Preference-drift ablation by dataset}
\label{tab:preference_drift_by_dataset}
\scriptsize
\setlength{\tabcolsep}{4.2pt}
\renewcommand{\arraystretch}{1.12}
\begin{adjustbox}{width=0.8\textwidth}
\begin{tabular}{llccccc}
\toprule
\textbf{Dataset} & \textbf{$\rho_\omega$}
& \textbf{FHV} $\uparrow$
& \textbf{FUT} $\uparrow$
& \makecell{\textbf{MC}\\$(\times 10^{-3})$ $\downarrow$}
& \textbf{SR} $\uparrow$
& \textbf{PCU} $\uparrow$ \\
\midrule

\multirow{4}{*}{MO-HalfCheetah-2d}
& 0.05
& $5.661\times 10^{6}$
& $2339.41$
& $0.899$
& $4.219$
& $4.146$ \\
& 0.10
& $5.787\times 10^{6}$
& $\mathbf{2341.87}$
& $0.826$
& $4.235$
& $4.194$ \\
& 0.20
& $\mathbf{5.790\times 10^{6}}$
& $2339.33$
& $\mathbf{0.818}$
& $4.405$
& $4.349$ \\
& 0.40
& $5.728\times 10^{6}$
& $2338.07$
& $0.875$
& $\mathbf{4.532}$
& $\mathbf{4.458}$ \\
\midrule

\multirow{4}{*}{MO-Ant-4d}
& 0.05
& $3.680\times 10^{14}$
& $4354.59$
& $1.641$
& $\mathbf{8.704}$
& $\mathbf{9.070}$ \\
& 0.10
& $3.407\times 10^{14}$
& $4271.80$
& $1.630$
& $8.607$
& $8.744$ \\
& 0.20
& $\mathbf{3.744\times 10^{14}}$
& $\mathbf{4369.02}$
& $\mathbf{1.572}$
& $8.699$
& $8.880$ \\
& 0.40
& $3.281\times 10^{14}$
& $4236.83$
& $1.752$
& $8.240$
& $8.253$ \\
\midrule

\multirow{4}{*}{MO-HalfCheetah-5d}
& 0.05
& $7.721\times 10^{16}$
& $2521.16$
& $1.663$
& $6.588$
& $5.442$ \\
& 0.10
& $9.184\times 10^{16}$
& $2662.65$
& $1.474$
& $\mathbf{6.797}$
& $\mathbf{7.346}$ \\
& 0.20
& $7.353\times 10^{16}$
& $2405.22$
& $1.477$
& $4.946$
& $4.347$ \\
& 0.40
& $\mathbf{9.484\times 10^{16}}$
& $\mathbf{2668.46}$
& $\mathbf{1.433}$
& $5.694$
& $6.199$ \\

\bottomrule
\end{tabular}
\end{adjustbox}
\end{table}

Table~\ref{tab:preference_drift_by_dataset} evaluates the sensitivity of
\algname{} to the preference drift rate $\rho_\omega$. The effect of
$\rho_\omega$ is dataset- and metric-dependent, reflecting a trade-off between
task diversity and history staleness. On MO-HalfCheetah-2d, moderate drift
$\rho_\omega=0.20$ gives the best FHV and lowest MC, while stronger drift
$\rho_\omega=0.40$ improves SR and PCU, suggesting that faster preference
changes mainly benefit online adaptation. On MO-Ant-4d, $\rho_\omega=0.20$
achieves the strongest FHV, FUT, and MC, whereas smaller drift
$\rho_\omega=0.05$ yields the best SR and PCU. On MO-HalfCheetah-5d, larger
drift $\rho_\omega=0.40$ gives the best feasible-front quality, while
$\rho_\omega=0.10$ provides the strongest online adaptation. Overall, these
results indicate that moderate-to-large preference drift can improve the
learning signal for task inference, but excessive or mismatched drift may reduce
post-change adaptation quality.

\subsection{Context-Dependent Dual Ablation}
\label{app:dual_ablation}


\begin{table}[!ht]
\centering
\caption{Context-dependent dual ablation}
\label{tab:ablation-dual-belief}
\scriptsize
\setlength{\tabcolsep}{4pt}
\renewcommand{\arraystretch}{1.12}
\begin{adjustbox}{width=0.9\textwidth}
\begin{tabular}{llcccccc}
\toprule
\textbf{Dataset} & \textbf{Variant}
& \textbf{FHV} $\uparrow$
& \textbf{FUT} $\uparrow$
& \textbf{SR} $\uparrow$
& \textbf{PCU} $\uparrow$
& \textbf{VR} $\downarrow$
& \textbf{MC} $\downarrow$ \\
\midrule

\multirow{4}{*}{MO-Ant-2d}
& $\lambda_\psi(\mathbf{z}_t)$
& $2.24{\times}10^{6}$ & $1.40{\times}10^{3}$ & 1.46 & 1.49 & 0 & $\mathbf{0.59{\times}10^{-3}}$ \\
& $\lambda_\psi(\widetilde B_t)$
& $5.76{\times}10^{6}$ & $2.22{\times}10^{3}$ & 3.53 & 3.51 & 0 & $1.21{\times}10^{-3}$ \\
& $\lambda$
& $5.78{\times}10^{6}$ & $2.22{\times}10^{3}$ & 4.06 & $\mathbf{4.19}$ & 0 & $1.49{\times}10^{-3}$ \\
& $\lambda_\psi(\widetilde B_t,\mathbf{z}_t)$
& $\mathbf{5.88{\times}10^{6}}$ & $\mathbf{2.25{\times}10^{3}}$ & $\mathbf{4.08}$ & 4.11 & 0 & $1.60{\times}10^{-3}$ \\
\midrule

\multirow{4}{*}{MO-HalfCheetah-2d}
& $\lambda_\psi(\mathbf{z}_t)$
& $2.33{\times}10^{6}$ & $1.51{\times}10^{3}$ & 2.09 & 2.11 & 0 & $\mathbf{0.45{\times}10^{-3}}$ \\
& $\lambda_\psi(\widetilde B_t)$
& $5.61{\times}10^{6}$ & $2.34{\times}10^{3}$ & 4.48 & 4.45 & 0 & $0.80{\times}10^{-3}$ \\
& $\lambda$
& $5.65{\times}10^{6}$ & $2.34{\times}10^{3}$ & 4.50 & 4.47 & 0 & $0.77{\times}10^{-3}$ \\
& $\lambda_\psi(\widetilde B_t,\mathbf{z}_t)$
& $\mathbf{5.66{\times}10^{6}}$ & $\mathbf{2.35{\times}10^{3}}$ & $\mathbf{4.53}$ & $\mathbf{4.52}$ & 0 & $0.79{\times}10^{-3}$ \\

\bottomrule
\end{tabular}
\end{adjustbox}
\end{table}

Table~\ref{tab:ablation-dual-belief} studies the role of the dual-variable
parameterization. The full context-dependent dual
$\lambda_\psi(\widetilde B_t,\mathbf{z}_t)$ achieves the strongest FHV, FUT, and
SR on both tasks, and also obtains the best PCU on MO-HalfCheetah-2d. Using only
the belief $\mathbf{z}_t$ produces the lowest MC, but substantially degrades
FHV, FUT, SR, and PCU, indicating overly conservative constraint enforcement.
The budget-only and global dual variants recover much of the performance, but
remain less consistent than conditioning jointly on the fused budget and belief.
These results support the use of a context-dependent dual variable that adapts
the constraint penalty to both feasibility level and latent task context.

\newpage
\section*{NeurIPS Paper Checklist}

\begin{enumerate}

\item {\bf Claims}
    \item[] Question: Do the main claims made in the abstract and introduction accurately reflect the paper's contributions and scope?
    \item[] Answer: \answerYes{}
    \item[] Justification: The abstract and Introduction state the main scope of the paper: constrained preference-conditioned MORL under drifting and partially observed preferences and budgets. The claims are supported by the proposed BECON framework, the theoretical error decomposition, and the empirical results in Sections 4-6.
    \item[] Guidelines:
    \begin{itemize}
        \item The answer \answerNA{} means that the abstract and introduction do not include the claims made in the paper.
        \item The abstract and/or introduction should clearly state the claims made, including the contributions made in the paper and important assumptions and limitations. A \answerNo{} or \answerNA{} answer to this question will not be perceived well by the reviewers. 
        \item The claims made should match theoretical and experimental results, and reflect how much the results can be expected to generalize to other settings. 
        \item It is fine to include aspirational goals as motivation as long as it is clear that these goals are not attained by the paper. 
    \end{itemize}

\item {\bf Limitations}
    \item[] Question: Does the paper discuss the limitations of the work performed by the authors?
    \item[] Answer: \answerYes{}
\item[] Justification: Section 7 discusses the key limitations,
including stage-wise constraints, latent-task identifiability,
and the scope of task drift.
    \item[] Guidelines:
    \begin{itemize}
        \item The answer \answerNA{} means that the paper has no limitation while the answer \answerNo{} means that the paper has limitations, but those are not discussed in the paper. 
        \item The authors are encouraged to create a separate ``Limitations'' section in their paper.
        \item The paper should point out any strong assumptions and how robust the results are to violations of these assumptions (e.g., independence assumptions, noiseless settings, model well-specification, asymptotic approximations only holding locally). The authors should reflect on how these assumptions might be violated in practice and what the implications would be.
        \item The authors should reflect on the scope of the claims made, e.g., if the approach was only tested on a few datasets or with a few runs. In general, empirical results often depend on implicit assumptions, which should be articulated.
        \item The authors should reflect on the factors that influence the performance of the approach. For example, a facial recognition algorithm may perform poorly when image resolution is low or images are taken in low lighting. Or a speech-to-text system might not be used reliably to provide closed captions for online lectures because it fails to handle technical jargon.
        \item The authors should discuss the computational efficiency of the proposed algorithms and how they scale with dataset size.
        \item If applicable, the authors should discuss possible limitations of their approach to address problems of privacy and fairness.
        \item While the authors might fear that complete honesty about limitations might be used by reviewers as grounds for rejection, a worse outcome might be that reviewers discover limitations that aren't acknowledged in the paper. The authors should use their best judgment and recognize that individual actions in favor of transparency play an important role in developing norms that preserve the integrity of the community. Reviewers will be specifically instructed to not penalize honesty concerning limitations.
    \end{itemize}

\item {\bf Theory assumptions and proofs}
    \item[] Question: For each theoretical result, does the paper provide the full set of assumptions and a complete (and correct) proof?
    \item[] Answer: \answerYes{}
\item[] Justification: Section 5 states the theoretical
assumptions, with complete proofs and auxiliary results
provided in Appendix C.
    \item[] Guidelines:
    \begin{itemize}
        \item The answer \answerNA{} means that the paper does not include theoretical results. 
        \item All the theorems, formulas, and proofs in the paper should be numbered and cross-referenced.
        \item All assumptions should be clearly stated or referenced in the statement of any theorems.
        \item The proofs can either appear in the main paper or the supplemental material, but if they appear in the supplemental material, the authors are encouraged to provide a short proof sketch to provide intuition. 
        \item Inversely, any informal proof provided in the core of the paper should be complemented by formal proofs provided in appendix or supplemental material.
        \item Theorems and Lemmas that the proof relies upon should be properly referenced. 
    \end{itemize}

    \item {\bf Experimental result reproducibility}
    \item[] Question: Does the paper fully disclose all the information needed to reproduce the main experimental results of the paper to the extent that it affects the main claims and/or conclusions of the paper (regardless of whether the code and data are provided or not)?
    \item[] Answer: \answerYes{}
  \item[] Justification: Section 6 and Appendices D--F
provide the benchmarks, training protocol, evaluation
metrics, implementation details, and additional analyses
needed for reproducibility.
    \item[] Guidelines:
    \begin{itemize}
        \item The answer \answerNA{} means that the paper does not include experiments.
        \item If the paper includes experiments, a \answerNo{} answer to this question will not be perceived well by the reviewers: Making the paper reproducible is important, regardless of whether the code and data are provided or not.
        \item If the contribution is a dataset and\slash or model, the authors should describe the steps taken to make their results reproducible or verifiable. 
        \item Depending on the contribution, reproducibility can be accomplished in various ways. For example, if the contribution is a novel architecture, describing the architecture fully might suffice, or if the contribution is a specific model and empirical evaluation, it may be necessary to either make it possible for others to replicate the model with the same dataset, or provide access to the model. In general. releasing code and data is often one good way to accomplish this, but reproducibility can also be provided via detailed instructions for how to replicate the results, access to a hosted model (e.g., in the case of a large language model), releasing of a model checkpoint, or other means that are appropriate to the research performed.
        \item While NeurIPS does not require releasing code, the conference does require all submissions to provide some reasonable avenue for reproducibility, which may depend on the nature of the contribution. For example
        \begin{enumerate}
            \item If the contribution is primarily a new algorithm, the paper should make it clear how to reproduce that algorithm.
            \item If the contribution is primarily a new model architecture, the paper should describe the architecture clearly and fully.
            \item If the contribution is a new model (e.g., a large language model), then there should either be a way to access this model for reproducing the results or a way to reproduce the model (e.g., with an open-source dataset or instructions for how to construct the dataset).
            \item We recognize that reproducibility may be tricky in some cases, in which case authors are welcome to describe the particular way they provide for reproducibility. In the case of closed-source models, it may be that access to the model is limited in some way (e.g., to registered users), but it should be possible for other researchers to have some path to reproducing or verifying the results.
        \end{enumerate}
    \end{itemize}

\item {\bf Open access to data and code}
    \item[] Question: Does the paper provide open access to the data and code, with sufficient instructions to faithfully reproduce the main experimental results, as described in supplemental material?
   \item[] Answer: \answerYes{}.
\item[] Justification: The code is publicly available, along with instructions for reproducing the experimental results.
    \item[] Guidelines:
    \begin{itemize}
        \item The answer \answerNA{} means that paper does not include experiments requiring code.
        \item Please see the NeurIPS code and data submission guidelines (\url{https://neurips.cc/public/guides/CodeSubmissionPolicy}) for more details.
        \item While we encourage the release of code and data, we understand that this might not be possible, so \answerNo{} is an acceptable answer. Papers cannot be rejected simply for not including code, unless this is central to the contribution (e.g., for a new open-source benchmark).
        \item The instructions should contain the exact command and environment needed to run to reproduce the results. See the NeurIPS code and data submission guidelines (\url{https://neurips.cc/public/guides/CodeSubmissionPolicy}) for more details.
        \item The authors should provide instructions on data access and preparation, including how to access the raw data, preprocessed data, intermediate data, and generated data, etc.
        \item The authors should provide scripts to reproduce all experimental results for the new proposed method and baselines. If only a subset of experiments are reproducible, they should state which ones are omitted from the script and why.
        \item At submission time, to preserve anonymity, the authors should release anonymized versions (if applicable).
        \item Providing as much information as possible in supplemental material (appended to the paper) is recommended, but including URLs to data and code is permitted.
    \end{itemize}

\item {\bf Experimental setting/details}
    \item[] Question: Does the paper specify all the training and test details (e.g., data splits, hyperparameters, how they were chosen, type of optimizer) necessary to understand the results?
    \item[] Answer: \answerYes{}
\item[] Justification: Section 6 specifies the experimental
protocol, with environment settings in Appendix D and
implementation details and hyperparameters in Appendix E.
    \item[] Guidelines:
    \begin{itemize}
        \item The answer \answerNA{} means that the paper does not include experiments.
        \item The experimental setting should be presented in the core of the paper to a level of detail that is necessary to appreciate the results and make sense of them.
        \item The full details can be provided either with the code, in appendix, or as supplemental material.
    \end{itemize}

\item {\bf Experiment statistical significance}
    \item[] Question: Does the paper report error bars suitably and correctly defined or other appropriate information about the statistical significance of the experiments?
    \item[] Answer: \answerYes{}
    \item[] Justification: The main experimental tables report mean and standard deviation over five independent runs, and Figure 2 includes uncertainty bands over seeds. These error bars capture variability due to random initialization and environment sampling.
    \item[] Guidelines:
    \begin{itemize}
        \item The answer \answerNA{} means that the paper does not include experiments.
        \item The authors should answer \answerYes{} if the results are accompanied by error bars, confidence intervals, or statistical significance tests, at least for the experiments that support the main claims of the paper.
        \item The factors of variability that the error bars are capturing should be clearly stated (for example, train/test split, initialization, random drawing of some parameter, or overall run with given experimental conditions).
        \item The method for calculating the error bars should be explained (closed form formula, call to a library function, bootstrap, etc.)
        \item The assumptions made should be given (e.g., Normally distributed errors).
        \item It should be clear whether the error bar is the standard deviation or the standard error of the mean.
        \item It is OK to report 1-sigma error bars, but one should state it. The authors should preferably report a 2-sigma error bar than state that they have a 96\% CI, if the hypothesis of Normality of errors is not verified.
        \item For asymmetric distributions, the authors should be careful not to show in tables or figures symmetric error bars that would yield results that are out of range (e.g., negative error rates).
        \item If error bars are reported in tables or plots, the authors should explain in the text how they were calculated and reference the corresponding figures or tables in the text.
    \end{itemize}

\item {\bf Experiments compute resources}
    \item[] Question: For each experiment, does the paper provide sufficient information on the computer resources (type of compute workers, memory, time of execution) needed to reproduce the experiments?
    \item[] Answer: \answerYes{}
    \item[] Justification: These details are provided in the appendix.
    \item[] Guidelines:
    \begin{itemize}
        \item The answer \answerNA{} means that the paper does not include experiments.
        \item The paper should indicate the type of compute workers CPU or GPU, internal cluster, or cloud provider, including relevant memory and storage.
        \item The paper should provide the amount of compute required for each of the individual experimental runs as well as estimate the total compute. 
        \item The paper should disclose whether the full research project required more compute than the experiments reported in the paper (e.g., preliminary or failed experiments that didn't make it into the paper). 
    \end{itemize}
    
\item {\bf Code of ethics}
    \item[] Question: Does the research conducted in the paper conform, in every respect, with the NeurIPS Code of Ethics \url{https://neurips.cc/public/EthicsGuidelines}?
    \item[] Answer: \answerYes{}
    \item[] Justification: The study is conducted as algorithmic research and conforms to the NeurIPS Code of Ethics.
    \item[] Guidelines:
    \begin{itemize}
        \item The answer \answerNA{} means that the authors have not reviewed the NeurIPS Code of Ethics.
        \item If the authors answer \answerNo, they should explain the special circumstances that require a deviation from the Code of Ethics.
        \item The authors should make sure to preserve anonymity (e.g., if there is a special consideration due to laws or regulations in their jurisdiction).
    \end{itemize}

\item {\bf Broader impacts}
    \item[] Question: Does the paper discuss both potential positive societal impacts and negative societal impacts of the work performed?
    \item[] Answer: \answerNA{}
    \item[] Justification: This work is primarily a methodological contribution to constrained multi-objective reinforcement learning. Its possible societal effects depend on how future systems instantiate and deploy such algorithms, rather than on the proposed method itself. Since the paper does not introduce a specific real-world deployment, user-facing system, or high-risk data/model release, we do not identify a direct broader impact requiring separate discussion.
    \item[] Guidelines:
    \begin{itemize}
        \item The answer \answerNA{} means that there is no societal impact of the work performed.
        \item If the authors answer \answerNA{} or \answerNo, they should explain why their work has no societal impact or why the paper does not address societal impact.
        \item Examples of negative societal impacts include potential malicious or unintended uses (e.g., disinformation, generating fake profiles, surveillance), fairness considerations (e.g., deployment of technologies that could make decisions that unfairly impact specific groups), privacy considerations, and security considerations.
        \item The conference expects that many papers will be foundational research and not tied to particular applications, let alone deployments. However, if there is a direct path to any negative applications, the authors should point it out. For example, it is legitimate to point out that an improvement in the quality of generative models could be used to generate Deepfakes for disinformation. On the other hand, it is not needed to point out that a generic algorithm for optimizing neural networks could enable people to train models that generate Deepfakes faster.
        \item The authors should consider possible harms that could arise when the technology is being used as intended and functioning correctly, harms that could arise when the technology is being used as intended but gives incorrect results, and harms following from (intentional or unintentional) misuse of the technology.
        \item If there are negative societal impacts, the authors could also discuss possible mitigation strategies (e.g., gated release of models, providing defenses in addition to attacks, mechanisms for monitoring misuse, mechanisms to monitor how a system learns from feedback over time, improving the efficiency and accessibility of ML).
    \end{itemize}
    
\item {\bf Safeguards}
    \item[] Question: Does the paper describe safeguards that have been put in place for responsible release of data or models that have a high risk for misuse (e.g., pre-trained language models, image generators, or scraped datasets)?
    \item[] Answer: \answerNA{}
    \item[] Justification: The paper does not release high-risk generative models, scraped datasets, or assets with substantial misuse risk.
    \item[] Guidelines:
    \begin{itemize}
        \item The answer \answerNA{} means that the paper poses no such risks.
        \item Released models that have a high risk for misuse or dual-use should be released with necessary safeguards to allow for controlled use of the model, for example by requiring that users adhere to usage guidelines or restrictions to access the model or implementing safety filters. 
        \item Datasets that have been scraped from the Internet could pose safety risks. The authors should describe how they avoided releasing unsafe images.
        \item We recognize that providing effective safeguards is challenging, and many papers do not require this, but we encourage authors to take this into account and make a best faith effort.
    \end{itemize}

\item {\bf Licenses for existing assets}
    \item[] Question: Are the creators or original owners of assets (e.g., code, data, models), used in the paper, properly credited and are the license and terms of use explicitly mentioned and properly respected?
    \item[] Answer: \answerYes{}
    \item[] Justification: The paper properly credits the creators and original owners of all existing benchmarks, baselines, and software assets used in the experiments. We respect the corresponding licenses and terms of use, with attribution and usage details provided in the paper and supplementary material.
    \item[] Guidelines:
    \begin{itemize}
        \item The answer \answerNA{} means that the paper does not use existing assets.
        \item The authors should cite the original paper that produced the code package or dataset.
        \item The authors should state which version of the asset is used and, if possible, include a URL.
        \item The name of the license (e.g., CC-BY 4.0) should be included for each asset.
        \item For scraped data from a particular source (e.g., website), the copyright and terms of service of that source should be provided.
        \item If assets are released, the license, copyright information, and terms of use in the package should be provided. For popular datasets, \url{paperswithcode.com/datasets} has curated licenses for some datasets. Their licensing guide can help determine the license of a dataset.
        \item For existing datasets that are re-packaged, both the original license and the license of the derived asset (if it has changed) should be provided.
        \item If this information is not available online, the authors are encouraged to reach out to the asset's creators.
    \end{itemize}

\item {\bf New assets}
    \item[] Question: Are new assets introduced in the paper well documented and is the documentation provided alongside the assets?
    \item[] Answer: \answerNA{}
    \item[] Justification: The paper does not introduce a new dataset, benchmark, or pretrained model as a primary asset.
    \item[] Guidelines:
    \begin{itemize}
        \item The answer \answerNA{} means that the paper does not release new assets.
        \item Researchers should communicate the details of the dataset\slash code\slash model as part of their submissions via structured templates. This includes details about training, license, limitations, etc. 
        \item The paper should discuss whether and how consent was obtained from people whose asset is used.
        \item At submission time, remember to anonymize your assets (if applicable). You can either create an anonymized URL or include an anonymized zip file.
    \end{itemize}

\item {\bf Crowdsourcing and research with human subjects}
    \item[] Question: For crowdsourcing experiments and research with human subjects, does the paper include the full text of instructions given to participants and screenshots, if applicable, as well as details about compensation (if any)? 
    \item[] Answer: \answerNA{}
    \item[] Justification: The paper does not involve crowdsourcing, user studies, or experiments with human subjects.
    \item[] Guidelines:
    \begin{itemize}
        \item The answer \answerNA{} means that the paper does not involve crowdsourcing nor research with human subjects.
        \item Including this information in the supplemental material is fine, but if the main contribution of the paper involves human subjects, then as much detail as possible should be included in the main paper. 
        \item According to the NeurIPS Code of Ethics, workers involved in data collection, curation, or other labor should be paid at least the minimum wage in the country of the data collector. 
    \end{itemize}

\item {\bf Institutional review board (IRB) approvals or equivalent for research with human subjects}
    \item[] Question: Does the paper describe potential risks incurred by study participants, whether such risks were disclosed to the subjects, and whether Institutional Review Board (IRB) approvals (or an equivalent approval/review based on the requirements of your country or institution) were obtained?
    \item[] Answer: \answerNA{}
    \item[] Justification: The paper does not involve human subjects, crowdsourcing, personal data collection, or user studies.
    \item[] Guidelines:
    \begin{itemize}
        \item The answer \answerNA{} means that the paper does not involve crowdsourcing nor research with human subjects.
        \item Depending on the country in which research is conducted, IRB approval (or equivalent) may be required for any human subjects research. If you obtained IRB approval, you should clearly state this in the paper. 
        \item We recognize that the procedures for this may vary significantly between institutions and locations, and we expect authors to adhere to the NeurIPS Code of Ethics and the guidelines for their institution. 
        \item For initial submissions, do not include any information that would break anonymity (if applicable), such as the institution conducting the review.
    \end{itemize}

\item {\bf Declaration of LLM usage}
    \item[] Question: Does the paper describe the usage of LLMs if it is an important, original, or non-standard component of the core methods in this research? Note that if the LLM is used only for writing, editing, or formatting purposes and does \emph{not} impact the core methodology, scientific rigor, or originality of the research, declaration is not required.
    \item[] Answer: \answerNA{}
    \item[] Justification: The proposed method does not use LLMs as an important, original, or non-standard component of the core methodology. 
    \item[] Guidelines:
    \begin{itemize}
        \item The answer \answerNA{} means that the core method development in this research does not involve LLMs as any important, original, or non-standard components.
        \item Please refer to our LLM policy in the NeurIPS handbook for what should or should not be described.
    \end{itemize}

\end{enumerate}

\end{document}